\documentclass[english]{IEEEtran}
\usepackage[T1]{fontenc}
\usepackage[latin9]{inputenc}
\usepackage{color}
\usepackage{babel}
\usepackage{array}
\usepackage{float}
\usepackage{multirow}
\usepackage{amsmath}
\usepackage{amssymb}
\usepackage{graphicx}
\usepackage{rotating}
\usepackage[numbers]{natbib}
\usepackage[unicode=true,pdfusetitle,
 bookmarks=true,bookmarksnumbered=false,bookmarksopen=false,
 breaklinks=false,pdfborder={0 0 0},pdfborderstyle={},backref=false,colorlinks=true]
 {hyperref}
\hypersetup{
 linkcolor=coolblack,citecolor=burntorange,urlcolor=burntgreen}

\makeatletter

\newcommand{\noun}[1]{\textsc{#1}}
\providecommand{\tabularnewline}{\\}
\floatstyle{ruled}
\newfloat{algorithm}{tbp}{loa}
\providecommand{\algorithmname}{Algorithm}
\floatname{algorithm}{\protect\algorithmname}

\usepackage{color}
\definecolor{burntorange}{rgb}{0.8, 0.33, 0.0}
\definecolor{charcoal}{rgb}{0.21, 0.27, 0.31}
\definecolor{coolblack}{rgb}{0.0, 0.28, 0.49}
\definecolor{burntgreen}{rgb}{0.05, 0.45, 0.27}
\definecolor{burntblue}{rgb}{0.05, 0.27, 0.8}

\usepackage{algpseudocode}
\usepackage{wasysym}
\usepackage{bbm}
\usepackage{breakurl}

\makeatother

\begin{document}
\title{Recombinator-$k$-means: An evolutionary algorithm that exploits $k$-means++
for recombination}
\author{Carlo Baldassi\\
carlo.baldassi@unibocconi.it\\
Artificial Intelligence Lab, Institute for Data Science and Analytics,
Bocconi University, via Sarfatti 25, 20135 Milan, Italy}
\maketitle
\begin{abstract}
We introduce an evolutionary algorithm called recombinator-$k$-means
for optimizing the highly non-convex kmeans problem. Its defining
feature is that its crossover step involves all the members of the
current generation, stochastically recombining them with a repurposed
variant of the $k$-means++ seeding algorithm. The recombination also
uses a reweighting mechanism that realizes a progressively sharper
stochastic selection policy and ensures that the population eventually
coalesces into a single solution. We compare this scheme with state-of-the-art
alternative, a more standard genetic algorithm with deterministic
pairwise-nearest-neighbor crossover and an elitist selection policy,
of which we also provide an augmented and efficient implementation.
Extensive tests on large and challenging datasets (both synthetic
and real-word) show that for fixed population sizes recombinator-$k$-means
is generally superior in terms of the optimization objective, at the
cost of a more expensive crossover step. When adjusting the population
sizes of the two algorithms to match their running times, we find
that for short times the (augmented) pairwise-nearest-neighbor method
is always superior, while at longer times recombinator-$k$-means
will match it and, on the most difficult examples, take over. We conclude
that the reweighted whole-population recombination is more costly,
but generally better at escaping local minima. Moreover, it is algorithmically
simpler and more general (it could be applied even to $k$-medians
or $k$-medoids, for example). Our implementations are publicly available
at \href{https://github.com/carlobaldassi/RecombinatorKMeans.jl}{https://github.com/carlobaldassi/RecombinatorKMeans.jl}.
\end{abstract}


\section{Introduction}

The problem of minimizing the sum-of-squares error ($\mathrm{SSE}$)
is a central and paradigmatic clustering problem in data science.
It is usually addressed by some variant of the classical $k$-means
heuristic \citep{wu2008top,berkhin2006survey}. The SSE objective
is most commonly defined as follows: given $N$ data points $\mathcal{X}=\left(x_{i}\right)_{i=1}^{N}$,
where each point is $D$-dimensional, $x_{i}\in\mathbb{R}^{D}$, and
given an integer $k\ge2$, we wish to find $k$ centroids $\mathcal{C}=\left(c_{a}\right)_{a=1}^{k}\in\left(\mathbb{R}^{D}\right)^{k}$,
and a partition of the data points $\mathcal{P}=\left(p_{i}\right)_{i=1}^{N}\in\left\{ 1,\dots,k\right\} ^{N}$
that minimize the cost function:
\begin{equation}
\mathrm{SSE}\left(\mathcal{C},\mathcal{P};\mathcal{X}\right)=\sum_{i=1}^{N}\left\Vert x_{i}-c_{p_{i}}\right\Vert ^{2}\label{eq:SSE}
\end{equation}

It is straightforward to prove that, for fixed $\mathcal{C}$, the
optimal partition is given by associating each point to its nearest
centroid; conversely, for fixed $\mathcal{P}$, the optimal centroids
are given by the barycenter of each cluster. Indeed, the standard
algorithm used for optimization is Lloyd's algorithm \citep{lloyd1982least},
commonly referred to simply as $k$-means, which, starting from an
initial guess for the centroids (the so-called ``seeding'') uses
an alternating iterative strategy, optimizing the partition and the
centroids in turn. This can be regarded as a local search strategy
and it is guaranteed to reach a fixed point in a finite number of
steps. However, the $\mathrm{SSE}$ function, when considered as a
function of $\mathcal{C}$ alone (with $\mathcal{P}$ optimized away
as explained above) is in general highly non-convex, and in fact the
optimization problem is NP-hard \citep{aloise2009np}. As a result,
$k$-means gets very easily stuck in local minima unless it is initialized
close to the global optimum.

Proposals about how to deal with this issue abound in the literature
\citep{celebi2013comparative,franti2019much}. The most direct approach
consists in trying to optimize the seeding. A basic (and computationally
very cheap) seeding strategy, which is often implied when referring
to ``the $k$-means algorithm'' without qualifiers, is to sample
the initial centroids uniformly at random from $\mathcal{X}$ \citep{macqueen1967some}.
This often leads to poor results and long convergence times. Alternative,
more refined schemes, e.g. $k$-means++ \citep{arthur2007k}, generally
trade off some computational time during the seeding procedure for
faster convergence times and improved results. However, it is often
the case under realistic circumstances that the optimization landscape
is so highly non-convex and riddled with local minima that local search
algorithms still yield very sub-optimal results, even with improved
seeding.

When looking to go beyond local search strategies, evolutionary algorithms
provide an appealing and successful general paradigm \citep{goldberg1989genetic,holland1992adaptation,koza1994genetic}.
Indeed, several approaches tailored to clustering problems, in which
the $\mathrm{SSE}$ cost is identified with the fitness, have been
put forward \citep{hruschka2009survey}. In this paper, we investigate
in detail two evolutionary algorithms: a novel one, called recombinator-$k$-means,
that we detail below, and a genetic algorithm proposed in ref.~\citep{franti2000genetic}
that we refer to as GA-$k$-means in this paper. They both follow
this scheme: we maintain a population of $J$ individuals, from which
a new generation is obtained by a crossover mechanism followed by
local optimization of each individual (via Lloyd's algorithm or any
equivalent scheme), until some convergence criterion for the whole
population is met. Because of the use of local optimization, this
scheme could be categorized as memetic, rather than genetic \citep{hruschka2009survey,moscato1992memetic,merz2002clustering}.

The most peculiar feature of recombinator-$k$-means is its multiple-parent
(in fact, population-wide) crossover mechanism, based on a repurposed
variant of the $k$-means++ seeding algorithm. The basic idea is to
pool together all the centroids of the current population, and to
use $k$-means++ to ``recombine'' them, letting it choose centroids
from different individuals. This produces new stochastic seeds that
are pieced together from locally-optimal configurations of the previous
generation, and which are then optimized locally again. On top of
this basic mechanism, we adopt a $\left(\mu+\lambda\right)$-ES selection
strategy for survival, ensuring that the average $\mathrm{SSE}$ of
the population decreases monotonically at each generation. Furthermore,
we employ a novel selection-for-variation strategy that gradually
and self-adaptively drives the recombination from an exploration phase
to an exploitation one. The combination of the two selection strategies
guarantees that the population eventually coalesces into a single
configuration, thereby providing a natural stopping criterion.

The main contributions of this work can be summarized as follows.
1) We present a new evolutionary algorithm for $\mathrm{SSE}$ optimization
with an original crossover step and an original selection-for-variation
policy; the former is easily generalizable to clustering problems
in the same family, while the latter could in principle be exported
easily to an even larger set of optimization problems. 2) We provide
an efficient implementation of GA-$k$-means, and demonstrate the
benefit of augmenting it with improved seeding for the generation
of its initial population. 3) We present results on large realistic
datasets in terms of the trade-off between optimization quality and
time, in the hope of offering practically relevant insights.

The rest of the paper is organized as follows. In sec.~\ref{sec:prior}
we set up the framework of our study in the context of prior literature.
In sec.~\ref{sec:PNN} we review the GA-$k$-means algorithm. In
sec.~\ref{sec:kmeans++} we review the $k$-means++ seeding procedure,
in particular an enhanced variant of it called greedy-kmeans++ that
is seldom discussed in the literature. In sec.~\ref{sec:recombinator}
we introduce our recombinator-$k$-means scheme. In sec.~\ref{sec:experiments}
we present and analyze detailed numerical results on several challenging
synthetic and non-synthetic datasets. Sec.~\ref{sec:discussion}
has a final discussion.

\section{Framework and prior literature\label{sec:prior}}

In the general context of clustering, optimizing the $\mathrm{SSE}$
objective can be taxonomized as partitional (i.e. the clusters are
non-overlapping) and non-hierarchical. It's a hard combinatorial optimization
problem with a wide range of applications, from data analysis to data
compression. Indeed, the literature on optimizing the $\mathrm{SSE}$
objective is vast and covers a large variety of circumstances. In
the present work, we do not consider any modified versions of the
objective function (e.g. outlier detection mechanisms as in refs.~\citep{zhang1996birch,gan2017k}).
In particular, we will always assume the desired number of clusters
$k$ to be part of the input to the problem.

We further focus our attention on situations in which local optimization
(one run of seeding+Lloyd's algorithm) is significantly sub-optimal
(i.e. it has a zero or very low chance of approximating well the global
optimum), and computational resources allow to perform local optimizations
repeatedly. We want to consider fairly large datasets that pose a
significant and challenging optimization problem; we will however
assume that the data fits into memory, otherwise even Lloyd's algorithm
would not be practical. For huge datasets that don't fit in memory,
a techniques such as BIRCH \citep{zhang1996birch}, that lossily compresses
the data by pre-clustering it as it reads it, could still be employed;
any standard $\mathrm{SSE}$ optimization technique could then be
applied to the compressed representation by just adding some weights
in eq.~(\ref{eq:SSE}). We also always assume the data to be in a
dense format, since sparse data generally requires specialized techniques,
e.g.~\citep{liu2017sparse}.

Under these circumstances, using an evolutionary strategy looks promising,
and indeed, as mentioned in the introduction, many proposals of this
kind have been put forward. A survey of such approaches can be found
for example in ref.~\citep{hruschka2009survey}. Most of the approaches
can be more specifically categorized as genetic algorithms, and basically
all of them use so-called cluster-oriented operators, meaning that
they are specialized to the task at hand rather than using a task-independent
meta-heuristic. They also all use a fixed population size across generations.
Their distinguishing characteristics can be broken down into which
specific choices were made for the key features and operators of simulated
genetic evolution: 1) representation of the information about an individual
(also called ``chromosomes'') 2) construction of the initial population
(generation $0$); 3) crossover (also called ``recombination'')
mechanism; 4) selection-for-variation policy (i.e.~the relative weight
of the parents in producing the offspring) and selection-for-survival
policy (i.e.~which individuals contribute to the next generation);
5) mutation step.

Some common traits emerge in many of these key features when surveying
$\mathrm{SSE}$-optimizing algorithms. In particular, as far as we
can tell: 1) The representation (chromosomes) uses either the centroids
$\mathcal{C}$ or the partition $\mathcal{P}$, or both. This is natural
since, as we have mentioned, either of the two can implicitly define
the other. 2) The starting population is always some variant of random
uniform sampling in some space. 3) The crossover step, when present,
always uses at most two parent individuals, producing either one or
two children. 4) The selection policies are most commonly, although
not always, uniform for variation and either elitist or proportional
to the fitness for selection. 5) The mutation step is more variable
(e.g. perturbation of the centroids with random noise, random swapping
of a centroid with a data point, reassigning a data point from one
cluster to a nearby one), sometimes it is absent entirely.

The state-of-the-art in this area is, to the best of our knowledge,
the genetic algorithm proposed by Fränti in ref.~\citep{franti2000genetic},
that we refer to as GA-$k$-means. In summary, this method uses the
full configuration $\left(\mathcal{C},\mathcal{P}\right)$ as representation,
its initialization is based on a partition generated by picking centroids
uniformly at random from the dataset, it employs an elitist policy
for survival and uniform for variation, the (optional) mutation step
is random swap of a centroid with a data point (after which the partition
must be updated). Finally and most importantly, it uses deterministic
pairwise-nearest-neighbor (PNN) crossover. We provide a detailed description
of this method in the next section. Another algorithm, called self-adaptive
genetic algorithm \citep{kivijarvi2003self}, purports to beat GA-$k$-means
by using a meta-optimization strategy: it considers an array of genetic
operators, assigns a combination of them to each individual, and applies
the genetic evolution to those as well. While interesting, this strategy
has an additional level of complexity and it is very expensive; the
authors tests run for $1000$ generations and several hours on average,
whereas we are interested in much shorter time scales (for the same
type of data). Furthermore, the tests are performed at a fixed amount
of generations, and the self-adaptive version takes between $1.5$
and $2$ times as much time. For these reasons, we don't consider
it in the present study.

When framed in the context of the existing literature, our recombinator-$k$-means
algorithm (that we describe in detail in sec.~\ref{sec:recombinator})
shares the same general structure, but it is peculiar in several respects.
Most prominently, the recombination/crossover mechanism (which uses
the centroids $\mathcal{C}$ as its chromosomes) involves the whole
population to generate each offspring, rather than just two parents.
While using multiple parents isn't unheard of in the context of genetic
programming/evolutionary methods (see e.g.~refs.~\citep{tsutsui1998study,eiben2003multiparent}),
it is nevertheless uncommon. Our selection-for-variation policy is
also novel and stands out from all other algorithms. Additionally,
the arguments at the basis of our recombination method assume that
the population consists of (nearly) local optima. For this reason,
our initial population is not sampled uniformly at random but the
result of local optimization, and we apply local optimization after
each crossover step as well. Therefore, recombinator-$k$-means can
be regarded as a memetic algorithm. The GA-$k$-means algorithm also
uses local optimization (although in the original implementation it
was capped at $2$ Lloyd's iterations), and could be more precisely
called memetic as well.

\section{Review of the genetic algorithm with deterministic PNN crossover\label{sec:PNN}}

The main feature of GA-$k$-means is the use of the deterministic
pairwise-nearest-neighbor (PNN) crossover, because it is particularly
well-suited for implementation efficiency. For any two individuals,
represented by their configurations $\left(\mathcal{C}_{1},\mathcal{P}_{1}\right)$
and $\left(\mathcal{C}_{2},\mathcal{P}_{2}\right)$, a new configuration
is produced as follows. First, the configurations are merged into
one with at most $2k$ centroids, obtained by $\mathcal{C}_{m}=\mathcal{C}_{1}\cup\mathcal{C}_{2}$;
the optimal partition $\mathcal{P}_{m}$ is also easily computed.
This step takes $O\left(kD+N\right)$ time. The second step is half
of a Lloyd iteration: the centroids are updated based on $\mathcal{P}_{m}$.
This takes $O\left(ND\right)$ time. The third step consists in iteratively
merging pairs of clusters until only $k$ clusters remain. Merging
two clusters means taking the union of the corresponding data points.
As it turns out, thanks to the fact that the $\mathrm{SSE}$ uses
the squared Euclidean distance, the centroid of the union is the average
of the two centroids, weighted by the partition sizes. The two clusters
to be merged at each turn are the ``nearest-neighbors'', meaning
those whose merging would increase the least the $\mathrm{SSE}$.
The cost increase of merging two clusters can also be computed solely
from their centroids and sizes, once again thanks to the properties
of the $\mathrm{SSE}$. After each cluster merge, the merging costs
of all other pairs need to be updated. By performing all the computations
on the centroids alone, without the need to look at the data, we can
save a factor $N$. Thus this third step requires $O\left(Dk^{2}\tau\right)$,
where $\tau$ is related to the time required for the updates after
each merge, and it is at most $O\left(k\right)$ but generally much
lower in practice. After the merge operations have brought down the
number of clusters to $k$, we still need a final step in which the
partition information is updated based on the new centroids, which
is the other half of a Lloyd iteration and takes $O\left(NDk\right)$.

Overall, the computational cost for producing an offspring from two
configurations is $O\left(Dk^{2}\tau+NDk\right)$. This must be repeated
$J$ times in order to produce the next generation.

The selection-for-survival mechanism of GA-$k$-means is elitist.
For a given $J$, define $J_{e}=\left\lceil \left(1+\sqrt{1+8J}\right)/2\right\rceil $.
In the elitist scheme, only the best $J_{e}$ members of the population
produce offspring, by pairwise mating. If $J$ is a triangular number,
then $J=J_{e}\left(J_{e}-1\right)/2$ and the new generation is obtained
from all possible pairs in the ``elite'' subset. The fact that none
of the two parents is favored in the crossover, and that all parents
in the elite subset have the same number of children, means that the
selection-for-variation policy is uniform. If $J$ is not a triangular
number, only the first $J$ pairs (in lexicographic order) out of
the elite subset are used, which introduces a slight bias towards
fitter parents. 

In ref.~\citep{franti2000genetic}, each new configuration is optimized
with at most $2$ additional Lloyd's iterations. We found that $10$
gives a better time-cost trade-off. Also, the first generation is
produced by choosing centroids uniformly at random from the data,
and computing the corresponding partition. In our numerical experiments
(see sec.~\ref{sec:experiments}) we will show that using a more
expensive greedy-$k$-means++ seeding (see next section) followed
by Lloyd's algorithm (at most $10$ iterations) also consistently
pays off in terms of the time-cost trade-off.

As mentioned above, genetic algorithms commonly employ a mutation
mechanism to increase the genetic variation in the population. One
such mechanism, based on randomly swapping a centroid for a data point,
was explored in ref.~\citep{franti2000genetic}, but the author recommends
against it for performance reasons.

As a further optimization, the tests and code of ref.~\citep{franti2000genetic}
use the technique introduced in ref.~\citep{kaukoranta1999reduced}
to speed up Lloyd iterations, which is quite effective in saving some
distance computations in the partition update after the first iteration,
at the cost of keeping track of which centroids remain unchanged between
one iteration and the next. We also use this technique throughout
the paper.

\section{Review of the (greedy)-$k$-means++ seeding procedure\label{sec:kmeans++}}

Arguably~\citep{franti2019much}, the most popular seeding method
among those that go beyond uniform sampling is currently the so-called
$k$-means++ method \citep{arthur2007k}, due to its simplicity (conceptual
and in terms of implementation), versatility, availability, relatively
low computational cost $O\left(NDk\right)$, and generally good performances,
especially if restarts are a viable option \citep{celebi2013comparative}.
  The procedure consists in sampling the centroids from the data
points $\mathcal{X}$ progressively: the first one is sampled uniformly
at random, while each new one is sampled with a probability proportional
to the squared-distance from the nearest among the already-chosen
centroids.\footnote{The choice of using the squared distance is tightly related to the
SSE objective \citep{arthur2007k}.} This procedure can be refined by adding an extra sampling step: whenever
a new centroid (besides the first one) needs to be selected, $s$
candidates are sampled, and the one that minimizes the cost function
computed with the \emph{current} number of centroids is kept. This
is called greedy-$k$-means++ by Celebi et al\@. in ref.~\citep{celebi2013comparative}
(where they recommend it as one of the optimal choices for practical
purposes when restarts are a viable option), but we were unable to
find an original source for this algorithm.\footnote{This variant is used by the scikit-learn Python library \citep{scikit-learn,scikit-learn-site}.
A comment in the code refers to it as a port from the implementation
of the original authors of $k$-means++, but the URL where that was
located is now inactive, and the results in the original paper do
not use it. In the literature only the basic variant is generally
reported; in ref.~\citep{celebi2013comparative} the citation is
again only to the original $k$-means++ publication.

} The details of the algorithm are provided as a special case of the
reservoir-kmeans++ scheme discussed later, see Algorithm~\ref{alg:reservoir-recombinator}.

The parameter $s$ determines the amount of extra sampling. For $s=1$
we recover the basic $k$-means++ algorithm. In all our tests, we
have used the default value used by the scikit-learn library, $s=\left\lfloor 2+\log k\right\rfloor $,
which seems to provide a good trade-off between the improvement in
the initial configuration and the extra computational time required.\footnote{In ref.~\citep{celebi2013comparative} it seems that $s=\log\left(k\right)$
was used; it's unclear if it was truncated or rounded.} The computational complexity of this algorithm is $O\left(ksND\right)$.

\section{The recombinator-$k$-means scheme\label{sec:recombinator}}

As stated in the introduction, the basic idea of recombinator-$k$-means
is to use greedy-$k$-means++ as a crossover step, applying it on
the centroids pooled from a population of local optima. This idea
stems from an empirical observation, which we summarize in the following.
In Appendix~\ref{appendix:case-study} we provide a detailed case-study
analysis that demonstrates the mechanism in detail. Consider a case
in which the data $\mathcal{X}$ is isotropically clustered around
some centroids and the clusters are all about the same size and shape,
and well separated from each other, and the number of clusters $k$
is known -- in other words, a case for which the global minimizer
of the $\mathrm{SSE}$ objective is very close to the ground truth.
Due to the highly non-convex nature of the $\mathrm{SSE}$, a single
run of a local optimization algorithm may still have a very low chance
of ``hitting'' on the ground truth, even under such favorable circumstances,
even if initialized with a good seeding technique such as greedy-$k$-means++
(we show this in sec.~\ref{subsec:First-batch}). In such context,
the basic strategy of just repeating the optimization a number of
times (known as ``repeated-$k$-means'' or ``multi-start-$k$-means''~\citep{franti2019much})
can be very inefficient. On the other hand, the sub-optimality of
each repeated try can often be ascribed to a few easily identifiable
and mostly uncorrelated ``mistakes''. Thus the centroids found in
$J$ ``unsuccessful'' individual runs, when pooled together, are
actually likely to be tightly clustered around the ground-truth centroids.

We can consider this pool of centroids as a new dataset that is mostly
concentrated around $k$ very tight clusters (around the ground-truth
centroids) with a few outliers (the ``mistakes'', e.g. points lying
between two ground-truth clusters). Running the greedy-$k$-means++
seeding on such dataset is very likely to only hit each and all of
the tight clusters and ignore the outliers, for two reasons: 1) the
centroids in the tight clusters have a higher chance of being selected,
simply because they are in the majority; 2) once a centroid in one
of the tight clusters is chosen, all the other centroids in the same
cluster have a near-zero chance of being subsequently chosen, due
to the $k$-means++ probability reweighting that goes like the square
of the distance. The use of $s>1$ candidates in greedy-$k$-means++
further increases the chances of success. The overall effect is that
of ``recombining'' the previous results and produce a nearly-optimal
configuration with good probability.

Indeed, in such (quite artificial) situations it is often the case
that a single recombination of very few results can solve the optimization
problem, as we show in sec.~\ref{subsec:First-batch}. Perhaps surprisingly,
this crossover step proves to be effective even in more realistic
scenarios, producing configurations that, once locally optimized,
are often significantly better than any of the input ones. If, when
repeating the procedure several times, several new alternative configurations
are produced, it is natural to consider iterating this process for
several generations until no further progress can be made.

\begin{algorithm*}
\begin{algorithmic}[1]
\Function{reservoir-kmeans++}{\textcolor{burntblue}{$\mathcal{Y}$, $w$,} $\mathcal{X}$, $k$, $s$}
	\State Initialize an empty list of centroids $\mathcal{C}$
	\State Sample the first centroid $c_1$ at random \textcolor{burntblue}{from $\mathcal{Y}$ with probability $P\left(y_i\right)\propto w_i$}, and append it to $\mathcal{C}$
	\For{$a = 2,\dots,k$}
		\State $\phi^\mathrm{best} \gets \infty$
		\For{$b = 1,\dots,s$}
			\State Sample candidate centroid $c^\mathrm{cand}$ \textcolor{burntblue}{from $\mathcal{Y}$ with probability $P\left(y_i\right)\propto w_i\min_{a^\prime<a}d\left(y_{i},c_{a^\prime}\right)^{2}$}
			\State $\mathcal{C}^\mathrm{cand} \gets $ append candidate $c^\mathrm{cand}$ to (a copy of) $\mathcal{C}$
			\State $\phi^\mathrm{cand} \gets \mathrm{SSE}\left(\mathcal{C}^\mathrm{cand}; \mathcal{X}\right)$
			\If {$\phi^\mathrm{cand} < \phi^\mathrm{best}$}
				\State $c^\mathrm{best} \gets c^\mathrm{cand}$
				\State $\phi^\mathrm{best} \gets \phi^\mathrm{cand}$
			\EndIf
		\EndFor
		\State $\mathcal{C} \gets$ append $c^\mathrm{best}$ to $\mathcal{C}$
	\EndFor
	\State \Return $\mathcal{C}$
\EndFunction
\end{algorithmic}\medskip{}

\begin{algorithmic}[1]
\Function{recombinator-kmeans}{$\mathcal{X}$, $k$, $J$, $s=\left\lfloor \log k+2\right\rfloor$, $\Delta\beta=0.1$, $t_\mathrm{max}=10$}
	\State $\mathcal{Y} \gets \mathcal{X}$
	\State Initialize $w$ to a uniform vector, of the same length as $\mathcal{Y}$
	\State $\beta \gets 0$
	\State $\phi^\mathrm{best} \gets \infty$
	\State Initialize an empty list of configurations $\mathcal{K}$ and of costs $\Phi$
	\Repeat
		\State Initialize an empty list of configurations $\mathcal{K}^\mathrm{new}$ and of costs $\Phi^\mathrm{new}$
		\For{$r = 1,\dots,J$}
			\State $\mathcal{C} \gets \textsc{reservoir-kmeans++}\left(\mathcal{Y}, w, \mathcal{X}, k, s\right)$
			\State $\mathcal{C} \gets$ Run Lloyd's algorithm on $\mathcal{X}$ with initial points $\mathcal{C}$ for at most $t_\mathrm{max}$ steps
			\State Compute $\phi=\mathrm{SSE}\left(\mathcal{C}; \mathcal{X}\right)$ and append it to $\Phi^\mathrm{new}$
			\State Append $\mathcal{C}$ to $\mathcal{K}^\mathrm{new}$
			\If {$\phi < \phi^\mathrm{best}$}
				\State $\phi^\mathrm{best} \gets \phi$
				\State $\mathcal{C}^{\mathrm{best}} \gets \mathcal{C}$
			\EndIf
		\EndFor
		\State $\Phi, \mathcal{K} \gets \textsc{keepbest}\left(\Phi\cup\Phi^\mathrm{new}, \mathcal{K}\cup\mathcal{K}^\mathrm{new}, J\right)$
		\State $\beta \gets \beta + \Delta\beta$
		\State $w \gets \textsc{weights}\left(\Phi, \beta\right)$
		\State $\mathcal{Y} \gets$ list of all the centroids in $\mathcal{K}$
	\Until {$\mathrm{mean}\left(\Phi\right) \approx \min\left(\Phi\right)$}
	\State \Return $\mathcal{C}^{\mathrm{best}}$
\EndFunction
\end{algorithmic}

\caption{\label{alg:reservoir-recombinator}\textbf{Top:} the \noun{reservoir-kmeans++}
recombination algorithm. We have highlighted the differences with
respect to greedy-kmeans++ in blue: the latter is recovered using
$\mathcal{Y}=\mathcal{X}$ as the reservoir and uniform weights $w$;
by also setting $s=1$ we recover the standard (non-greedy) kmeans++
algorithm. \protect \\
\textbf{Bottom:} the \noun{recombinator-kmeans} algorithm. The use
of an approximate comparison in the stopping criterion (line 23) is
intended to account for small (arguably irrelevant) differences. In
our tests, we used a relative tolerance of $10^{-4}$. The functions
\noun{keepbest} and \textsc{weights} are discussed in the text. The
default values for $s$, $\Delta\beta$ and $t_{\max}$, used throughout
the paper, are shown in the argument list.}
\end{algorithm*}

This basic idea can be improved by introducing a few adjustments to
the greedy-$k$-means++ seeding algorithm to repurpose it for recombination;
we call the resulting algorithm reservoir-$k$-means++, see the first
function in Algorithm~\ref{alg:reservoir-recombinator}. It accepts
a reservoir argument $\mathcal{Y}$: a list of points from which to
sample the seeds, distinct from the data points $\mathcal{X}$. We
still use the original data $\mathcal{X}$ in the cost function $\mathrm{SSE}$
to determine the best candidate among the $s$ samples at each step.
There is one more extra argument, discussed below: a vector of weights
$w$ of the same size as $\mathcal{Y}$, that can be used to favor
some of the candidates in the reservoir over others, realizing a soft
selection-for-variation policy. The computations can be carried out
rather more efficiently than shown in the pseudocode, by employing
appropriate data structures. For the greedy-kmeans++ special case
(i.e., when $\mathcal{Y}=\mathcal{X}$), time scales as $O\left(kNDs\right)$,
with $O\left(ND\right)$ additional memory. Using a reservoir with
$kJ$ points instead we get $O\left(k\left(sN+kJ\right)D\right)$
time, with $O\left(kJD\right)$ extra memory. The two cases are comparable
as long as $kJ$ is $O\left(N\right)$. The actual code is publicly
available at ref.~\citep{RECcode}.

The second function in Algorithm.~\ref{alg:reservoir-recombinator}
is the \noun{recombinator-kmeans} algorithm. It uses two additional
auxiliary functions, \noun{keepbest} and \noun{weights}. The function
$\textsc{keepbest}$$\left(\Phi^{\prime},\mathcal{K}^{\prime},J\right)$
returns the $J$ best costs in $\Phi^{\prime}$, along with their
corresponding configurations found in $\mathcal{K}^{\prime}$. It
is, in fact, a realization of the so-called $\left(\mu+\lambda\right)$
selection-for-survival procedure used in evolutionary strategies \citep{back2018evolutionary}.
It guarantees that the population average monotonically improves at
each generation.

The function \textsc{weights} determines the selection-for-variation
bias that favors the centroids in $\mathcal{Y}$ belonging to configurations
of smaller cost; it thus has the role of a fitness function. We introduce
a novel heuristic for this purpose, that gradually increases the bias,
starting from a low value to encourage initial exploration and variability,
and progressively going towards an exploitation phase, ensuring that
the batches eventually collapse onto some ``consensus'' configuration.
In detail: given an array of costs $\Phi=\left(\phi_{a}\right)_{a=1}^{J}$,
we compute the best $\phi^{\star}=\min_{a=1:J}\left(\phi_{a}\right)$
and the mean $\overline{\phi}=\frac{1}{J}\sum_{a=1}^{J}\phi_{a}$,
and use the following formula to determine the weight of the centroids
in a sample $a$:\footnote{All the centroids in the sample receive the same weight, thus each
value $w_{a}$ gets repeated $k$ times when building the array that
is actually passed to \noun{reservoir-kmeans++}.}
\begin{equation}
w_{a}=\exp\left(-\beta\frac{\phi_{a}-\phi^{\star}}{\overline{\phi}-\phi^{\star}}\right)
\end{equation}
with some parameter $\beta\ge0$. This formula gives the largest weight
to configurations close to the best one. Using the difference between
the mean and the best as a scale in the denominator makes the function
self-adaptive. The parameter $\beta$ determines the amount of skew
in the weights, with $\beta=0$ corresponding to the flat (unweighted)
case, and $\beta\to\infty$ to only choosing the candidates from the
best configurations; we increase it by a constant amount $\Delta\beta$
at each iteration. The algorithm is not very sensitive to the precise
value of $\Delta\beta$, as long as it is non-zero; $\Delta\beta=0.1$
seems to be a good default value, and no extra tuning is required.
A further discussion of the effect of $\Delta\beta$, with some numerical
results, can be found in Appendix~\ref{appendix:parameters-deltab}.

The maximum number of Lloyd's iterations $t_{\mathrm{max}}$ was set
to $10$ in all our tests, which seems to provide a good trade-off
between computational time and optimization quality (see the Appendix~\ref{appendix:parameters-tmax}
for numerical results on this). Analogously to GA-$k$-means, we don't
need to completely reach convergence in a single run since, empirically,
$10$ Lloyd's iterations are normally enough to get at least close
to a fixed point, and if it is a good one then it will likely be (fully
or partially) picked up in the next round and optimized further. By
the point when convergence is achieved it's extremely likely that
the final configuration has been fully optimized. We confirmed that
this is indeed the case in all our tests. Therefore the only parameter
that needs to be tuned is $J$.

We also note that, in the limiting case when $\Delta\beta\to\infty$
and $t_{\mathrm{max}}=\infty$, \noun{recombinator-kmeans} emulates
multi-start-$k$-means, since it just performs $J$ independent local
optimizations, then it repeatedly selects the best one at the following
generation, and, having converged, stops.

\section{Numerical experiments\label{sec:experiments}}

\subsection{Experimental setup}

We performed a series of tests comparing \textsc{recombinator-kmeans}
with the GA-$k$-means algorithm discussed in sec.~\ref{sec:PNN}.
We refer to our own implementation as \noun{ga-kmeans}. We tested
two methods for initializing the population in \noun{ga-kmeans}: the
first follows ref.~\citep{franti2000genetic} and thus uses as initial
centroids $k$ points chosen uniformly at random from the dataset.
The partition is deduced from the centroids but no local optimization
is performed. We call this variant \noun{ga-kmeans-raw}. An augmented
variant, that we call \noun{ga-kmeans++}, uses greedy-$k$-means++
followed by Lloyd's algorithm (at most $10$ iterations) to initialize
the population, analogously to \textsc{recombinator-kmeans}.

Despite their similar structure, comparing the three algorithms is
not straightforward. For example, performing $J$ pairwise crossovers
is generally faster than performing $J$ recombinations, each of which
involves the whole population, but the latter may achieve results
that would require a much larger population with the former technique.
The number of Lloyd iterations alone does not take into account that
they become computationally cheaper near convergence, due to the optimizations
mentioned at the end of sec.~\ref{sec:PNN}. The additional seeding
effort of \noun{ga-kmeans++} compared to \noun{ga-kmeans-raw} may
be compensated by a reduced number of generations, etc.

We thus performed most of our comparisons, especially with realistic
data, in terms of the value of the $\mathrm{SSE}$ obtained as a function
of the wall-clock computational time. We performed all tests sequentially
with no other computationally intensive processes running while testing,
on the same hardware (Intel Core i7-9750H 2.60GHz CPU, 64Gb DDR4 2666MHz
RAM, running Ubuntu Linux 18.04 with 5.3.0 kernel); We used the same
programming language (Julia v1.6.2) for all our tests, in order to
allow as much sharing as possible of the code and data structures
and only highlight the algorithmic differences. To this end, we have
rewritten GA-$k$-means from scratch, following ref.~\citep{franti2000genetic}
and the C implementation obtained from ref.~\citep{GAcode}. Our
version implements the same basic algorithms (and can thus reproduce
the results) of the C code, but it is more optimized (mainly for cache
locality) and considerably faster (a factor of 2 or more, depending
on the dataset).

A few minor modifications in our implementation of the algorithm are:
1) Originally, the stopping criterion was to check if no improvement
was made in the last generation, but on top of this we also check
for population collapse, as for \noun{recombinator-kmeans} (cf.~Alg.~\ref{alg:reservoir-recombinator}),
which can save a generation occasionally. 2) We keep a separate record
of the best configuration seen so far, rather than including it in
the next generation by default as in the original C implementation;
this means that we are effectively using 1 more individual in the
population. 3) We do not preprocess the data, whereas the original
code scales each dimension individually.

After extensive preliminary testing, we used the following settings
in all our tests. We set a relative tolerance of $10^{-5}$ on the
cost for the convergence of Lloyd's algorithm, and a maximum of $t_{\mathrm{max}}=10$
iterations. We set a relative tolerance of $10^{-4}$ for population
collapse. For \textsc{recombinator-kmeans}, we set $s=\left\lfloor 2+\log k\right\rfloor $,
$\Delta\beta=0.1$. In all the algorithms, the only parameter that
we vary is thus $J$.

We have divided the tests in two batches. The first one consists of
synthetic datasets for which the correct $k$ is known and the optimal
$\mathrm{SSE}$ is close to the ground truth. This is mainly intended
to measure the ability of the algorithms to find the solution when
one can be clearly identified. The second one consists of challenging
real-world datasets, and in that case we simply measured how much
we could optimize the $\mathrm{SSE}$ as a function of the time spent
doing so. We also monitored the quality of the resulting clustering
compared to the ground truth, where available.

For the second batch of tests, we also compared the evolutionary algorithms
with a state-of-the-art non-evolutionary algorithm, called random
swap, which was proposed in ref.~\citep{franti2018efficiency}. It
consists in attempting random swaps between a centroid and a random
data point, followed by $t_{\mathrm{max}}=2$ Lloyd's iterations;
the swap is greedily accepted if it improves the $\mathrm{SSE}$ cost,
otherwise it is rejected and another swap is attempted. It does not
have a well-defined stopping criterion; in our implementation, we
used a wall-clock time limit. We used the same programming language
and data structures for this algorithm as well (the resulting code's
speed is comparable to or better than the C one from ref.~\citep{GAcode}),
and performed the tests under the same conditions as for the others.
Our implementation also improves on the original in two ways: 1) Once
the timer expires, we perform a final optimization with Lloyd's algorithm
and $t_{\mathrm{max}}=\infty$, using the same relative tolerance
$10^{-5}$ as for the other algorithms to detect convergence; 2) We
use greedy-$k$-means++ seeding to initialize the algorithm. In the
following, we refer to this algorithm as \noun{randswap-kmeans++}.

\begin{table}
\centering{}\caption{\label{tab:datasets}Characteristics of the datasets used in the tests.}
{\scriptsize{}}%
\begin{tabular}{|cc|ccc|}
\hline 
\multicolumn{2}{|c|}{{\scriptsize{}dataset}} & {\scriptsize{}$D$} & {\scriptsize{}$N$} & {\scriptsize{}$k$}\tabularnewline
\hline 
\hline 
\multirow{5}{*}{\begin{turn}{90}
{\scriptsize{}synthetic}
\end{turn}} & \emph{\scriptsize{}A3} & {\scriptsize{}$2$} & {\scriptsize{}$7500$} & {\scriptsize{}$50$}\tabularnewline
 & \emph{\scriptsize{}Birch1} & {\scriptsize{}$2$} & {\scriptsize{}$100000$} & {\scriptsize{}$100$}\tabularnewline
 & \emph{\scriptsize{}Birch2} & {\scriptsize{}$2$} & {\scriptsize{}$100000$} & {\scriptsize{}$100$}\tabularnewline
 & \emph{\scriptsize{}Unbalance} & {\scriptsize{}$2$} & {\scriptsize{}$6500$} & {\scriptsize{}$8$}\tabularnewline
 & \emph{\scriptsize{}Dim1024} & {\scriptsize{}$1024$} & {\scriptsize{}$1024$} & {\scriptsize{}$16$}\tabularnewline
\hline 
\hline 
\multirow{5}{*}{\begin{turn}{90}
{\scriptsize{}real-world}
\end{turn}} & \emph{\scriptsize{}Bridge} & {\scriptsize{}16} & {\scriptsize{}$4096$} & {\scriptsize{}$256$}\tabularnewline
 & \emph{\scriptsize{}House} & {\scriptsize{}$3$} & {\scriptsize{}$34112$} & {\scriptsize{}$256$}\tabularnewline
 & \emph{\scriptsize{}Miss America} & {\scriptsize{}$16$} & {\scriptsize{}$6480$} & {\scriptsize{}$256$}\tabularnewline
 & \emph{\scriptsize{}UrbanGB} & {\scriptsize{}$2$} & {\scriptsize{}$360177$} & {\scriptsize{}$469$}\tabularnewline
 & \emph{\scriptsize{}Olivetti} & {\scriptsize{}$4096$} & {\scriptsize{}$400$} & {\scriptsize{}$40$}\tabularnewline
\hline 
\end{tabular}
\end{table}

\subsection{Tests on synthetic datasets\label{subsec:First-batch}}

For the first batch of tests, we used five synthetic datasets from
from the repository of ref.~\citep{franti2018k}: \emph{A3}, \emph{Birch1},
\emph{Birch2}, \emph{Unbalance} and \emph{Dim1024}. Their characteristics
are summarized in the first part of table~\ref{tab:datasets}. We
uniformly scaled all the datasets by dividing their entries by the
overall maximum. They are all composed of fairly well-separated clusters,
in two dimensions (except \emph{Dim1024} which is very high-dimensional),
with data generated with isotropic Gaussians around known ground-truth
centroids, so that in practice optimizing the $\mathrm{SSE}$ objective
basically recovers the ground truth, and nevertheless escaping local
minima in the optimization process is not trivial.

For these datasets, a useful measure of the quality of the clustering
is given by the asymmetric centroid index (CI) as defined in ref.~\citep{franti2014centroid},
which counts the number of unmatched (``orphan'') ground-truth centroids
in the resulting clustering. Empirically, it is easy to observe that
the $\mathrm{SSE}$ values of the local minima close to the ground
truth for these datasets are separated in tight bands, each corresponding
to $\mathrm{CI=0,1,2,\dots}$ (an example for \emph{A3} is shown in
the Appendix, fig.~\ref{fig:hist-A3}). We can thus define the success
rate of an algorithm as the probability to obtain $\mathrm{CI=0}$.

The difficulty for \emph{A3}, \emph{Birch1} and \emph{Birch2} mostly
relies in the fact that $k$ is rather large ($50$ or $100$), whereas
the \emph{Unbalance} dataset has only $8$ very well-separated clusters,
but as the name implies they are very inhomogeneous (there are $3$
very dense clusters with $2000$ points each on one side and $5$
tiny clusters with $100$ points each on the other) which has a particularly
daunting effect on seeding algorithms that sample points uniformly
(since it's easy to miss at least some of the small clusters). Similarly
to \emph{Unbalance}, for \emph{Dim1024} the difficulty is in the large
separation between clusters which tends to trap Lloyd's algorithm
if the seeding misses some cluster; we also chose it to explore the
high-dimensional regime.

\begin{table*}
\begin{centering}
\caption{\label{tab:results1}Synthetic datasets: success rates and convergence
times}
{\scriptsize{}}%
\begin{tabular}{|c|cc|cc|cc|c|c|}
\hline 
\multirow{2}{*}{{\scriptsize{}dataset}} & \multicolumn{2}{c|}{\noun{\scriptsize{}recombinator}} & \multicolumn{2}{c|}{\noun{\scriptsize{}ga-raw}} & \multicolumn{2}{c|}{\noun{\scriptsize{}ga++}} & \noun{\scriptsize{}kmeans++} & \noun{\scriptsize{}kmeans}\tabularnewline
 & {\scriptsize{}succ.rate} & {\scriptsize{}time} & {\scriptsize{}succ.rate} & {\scriptsize{}time} & {\scriptsize{}succ.rate} & {\scriptsize{}time} & {\scriptsize{}succ.rate} & {\scriptsize{}succ.rate}\tabularnewline
\hline 
\hline 
\emph{\scriptsize{}A3} & {\scriptsize{}$100\%$} & {\scriptsize{}$0.089\pm0.014$} & {\scriptsize{}$99.5\%$} & {\scriptsize{}$0.031\pm0.003$} & {\scriptsize{}$100\%$} & {\scriptsize{}$0.058\pm0.003$} & {\scriptsize{}$5.4\%$} & {\scriptsize{}$0\%$}\tabularnewline
\emph{\scriptsize{}Birch1} & {\scriptsize{}$100\%$} & {\scriptsize{}$4.0\pm0.3$} & {\scriptsize{}$100\%$} & {\scriptsize{}$1.35\pm0.04$} & {\scriptsize{}$100\%$} & {\scriptsize{}$2.22\pm0.05$} & {\scriptsize{}$0.3\%$} & {\scriptsize{}$0\%$}\tabularnewline
\emph{\scriptsize{}Birch2} & {\scriptsize{}$100\%$} & {\scriptsize{}$2.2\pm0.4$} & {\scriptsize{}$100\%$} & {\scriptsize{}$0.61\pm0.05$} & {\scriptsize{}$100\%$} & {\scriptsize{}$1.27\pm0.06$} & {\scriptsize{}$7.6\%$} & {\scriptsize{}$0\%$}\tabularnewline
\emph{\scriptsize{}Unbalance} & {\scriptsize{}$100\%$} & {\scriptsize{}$\left(6\pm3\right)\cdot10^{-3}$} & {\scriptsize{}$67.2\%$} & {\scriptsize{}$\left(18\pm6\right)\cdot10^{-3}$} & {\scriptsize{}$100\%$} & {\scriptsize{}$\left(6\pm3\right)\cdot10^{-3}$} & {\scriptsize{}$94.6\%$} & {\scriptsize{}$0.01\%$}\tabularnewline
\emph{\scriptsize{}Dim1024} & {\scriptsize{}$100\%$} & {\scriptsize{}$0.082\pm0.04$} & {\scriptsize{}$99.5\%$} & {\scriptsize{}$0.043\pm0.006$} & {\scriptsize{}$100\%$} & {\scriptsize{}$0.088\pm0.002$} & {\scriptsize{}$100\%$} & {\scriptsize{}$0.02\%$}\tabularnewline
\hline 
\end{tabular}{\scriptsize\par}
\par\end{centering}
\begin{centering}
\par\end{centering}
\centering{}\medskip{}
Success rate (probability of finding a configuration with $\mathrm{CI}=0$)
and convergence time (mean and standard deviation, in seconds) for
different algorithms. The last two columns show the success rate of
a single run of Lloyd's algorithm, respectively with greedy-$k$-means++
or random uniform seeding. At least $100$ tests were performed for
each dataset and algorithm.
\end{table*}

For these datasets, we have used a small population of $J=5$, and
measured the success rate ($k$ was set to the ground truth value)
and the convergence time. We have also measured the success rate of
single runs of the Lloyd's algorithm, both with random-uniform and
with greedy-$k$-means++ seeding, for reference. The results are shown
in table~\ref{tab:results1}. Both \noun{recombinator-kmeans} and
\noun{ga-kmeans++} have 100\% success rate in all cases (and solve
the problem in fewer than $2$ generations on average in all cases);
between the two, \noun{ga-kmeans++} is faster. On the other hand,
\noun{ga-kmeans-raw} occasionally fails on \emph{A3} and \emph{Dim1024},
and is even worse than single-run with greedy-$k$-means++ initialization
for \emph{Unbalance}. In fact, \emph{Unbalance} and \emph{Dim1024}
are quite hard for uniform initialization, just like the rest of the
datasets (see last column), but they are easily solved by good seeding.
The other datasets instead are still rather hard for purely local
search even with greedy-$k$-means++ seeding, but they become quite
easy for population algorithms.

The \emph{Unbalance} result demonstrates that, even for GA-$k$-means,
good seeding may be crucial: if the initial population of centroids
misses some small but well-defined cluster, the crossover algorithm
alone is unable to correct the mistake. This issue in not tightly
related to the specific crossover function; it might be fixed in general
by some mutation mechanism, but that would introduce additional complexity
and require tuning. Some attempts to use the mutation mechanism proposed
in ref.~\citep{franti2000genetic} produced only minor improvements.
Using greedy-$k$-means++ seeding seems to be a better strategy, since
it is specifically designed to cover the dataset well, and furthermore
in difficult cases it's still able to introduce significant variability
in the initial population due to its stochastic nature.

With good seeding, both population algorithms are quite successful
at overcoming the cost barriers and recovering the ground truth. For
\noun{recombinator-kmeans}, in particular, the results support the
qualitative arguments of sec.~\ref{sec:recombinator} that inspired
the algorithm: for \emph{A3}, \emph{Birch1}, \emph{Birch2} the \noun{reservoir-kmeans++}
crossover can piece together the solution in very few generations
even with a small population. It is, however, more expensive than
PNN crossover, which works equally well in these synthetic scenarios.
A few additional tests (not reported here for brevity) performed on
non-synthetic datasets with small $k$ (but fairly large $N$ or $D$),
for which several algorithms can (arguably) find the global optimum
relatively easily, corroborate this scenario, including the crucial
role of seeding in some cases.

The next section explores cases in which instead local optimization
by itself would not typically get near the global optimum at all,
even with greedy-$k$-means++ seeding, and show that the recombination
crossover is still very effective, and can even become more convenient
in practice than the computationally cheaper PNN.

\subsection{Tests on real-world datasets\label{subsec:Second-batch}}

In the second batch of experiments, we tested five real-world datasets,
whose characteristics are summarized in the second part of table~\ref{tab:datasets}.
The first three are those that were also used as benchmarks in refs\@.~\citep{franti2000genetic,franti2018efficiency},
all downloaded from the repository of ref.~\citep{franti2018k}:
the \emph{Bridge} dataset (``non-binarized'' version), the \emph{House}
dataset (``8 bits per color'' version) and the first \emph{Miss
America} dataset (``frame 1 vs 2'' version). Contrary to the previous
batch, none of them appears to present a cluster structure that could
emerge by optimizing the $\mathrm{SSE}$ objective, and there is no
ground truth available; thus, we don't have a sharp notion of ``successful
clustering'' for these cases. The last two are: \emph{UrbanGB}, a
large dataset consisting of geographical coordinates of car accidents
occurred in urban areas within Great Britain that we have prepared
ourselves and is available at ref.~\citep{UCI-MachineLearningRepository}
(the details of how it was constructed are provided as metadata in
the repository); \emph{Olivetti}, a high-dimensional dataset from
AT\&T Laboratories Cambridge available via scikit-learn\footnote{Details available at ref.~\citep{OlivettiData}. The page contains
a link to the original page on the University of Cambridge website,
but that page no longer exists. We couldn't find a page for this dataset
on the University website.} \citep{scikit-learn,scikit-learn-site} consisting of grayscale images
of the faces of $40$ subjects in $10$ different poses each. We performed
no manipulations or normalizations on the data, except for \emph{UrbanGB}
where we scaled down the first dimension (longitude) by a factor or
$1.7$ to make the distance computations roughly reflect geographical
distances\footnote{A set of tests on unscaled data produced qualitatively analogous results.}. 

The \emph{UrbanGB} dataset is intended as an extremely challenging
version of the synthetic data analyzed in the previous batch: it comprises
a large number of blob-like clusters (the urban centers), which however
are not well separated, and exhibit significant heterogeneity, imbalance
and spatial non-uniformity. The \emph{Olivetti} dataset was chosen
for its very high dimensionality and for being computationally challenging;
it is also a case for which $\mathrm{SSE}$ optimization is a reasonably
realistic strategy to extract information (albeit certainly primitive
when used on raw data in the context of computer vision), since the
dimensions are in principle homogeneous, the images are centered,
and the clusters are balanced.

For the \emph{UrbanGB} and \emph{Olivetti} datasets a ground truth
is given, as a partition of the points, although it doesn't correspond
to a minimum of the $\mathrm{SSE}$. We have used the ground-truth
value of $k$ in our tests. We have also evaluated the quality of
the results by measuring the variation of information ($\mathrm{VI}$)
between the partitions \citep{meilua2007comparing}.

All of these datasets provide a quite hard optimization challenge.
Thus, we varied the population size $J$ of the evolutionary algorithms
and measured the average value of the objective $\mathrm{SSE}$ and
the convergence time, over $30$ or more independent runs of each
algorithm. As expected, a larger population leads to a better $\mathrm{SSE}$
but increases the convergence time. We did not use the same value
of $J$ for all algorithms, because for a given population size \noun{recombinator-kmeans}
systematically finds better $\mathrm{SSE}$ values, but \noun{ga-kmeans}
is much faster. In order to have comparable times between the two
family of algorithms, we proceeded as follows. For \noun{recombinator-kmeans},
we used $J\in\left\{ 5,10,20,40,80\right\} $ (for \emph{UrbanGB}
only $\left\{ 5,10,20\right\} $); for \noun{ga-kmeans++}, we looked
for a set of values of $J$ that would produce comparable convergence
times, but we restricted the search to triangular numbers so that
the entire ``elite'' part of the population could be exploited.
For \noun{ga-kmeans-raw} we just used the same set of $J$ as for
\noun{ga-kmeans++}, as the timing differences between the two are
generally not very large.

After all the tests with the evolutionary algorithms were completed,
we performed $10$ independent runs of \noun{randswap-kmeans++} for
each dataset, using the maximum average time of the evolutionary algorithms
as a time limit.

\begin{figure*}
\begin{centering}
\includegraphics[width=0.87\textwidth]{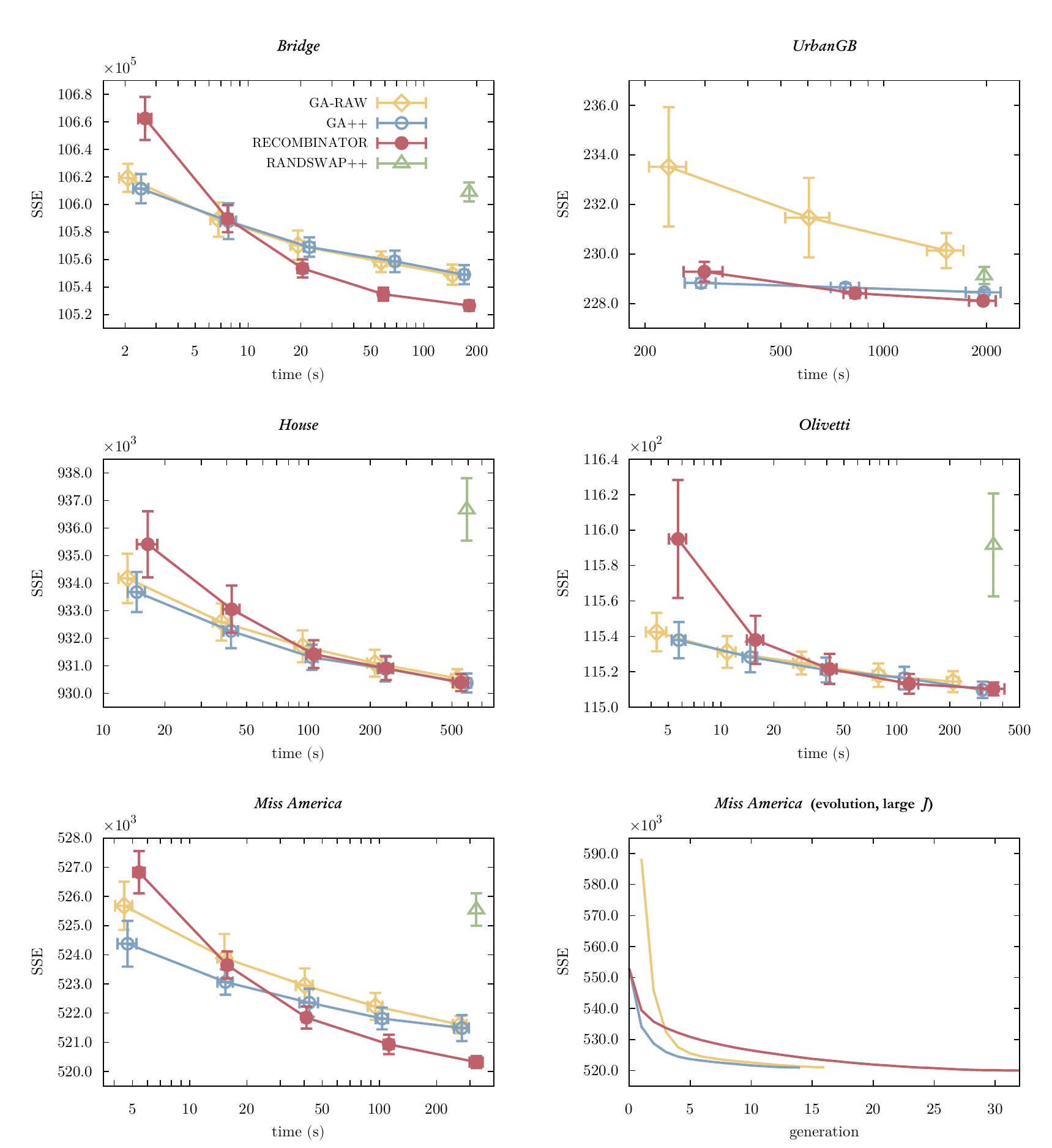}
\par\end{centering}
\caption{\label{fig:results2}Results for the tests on real-world datasets
(see table~\ref{tab:datasets}). First five panels: $\mathrm{SSE}$
cost vs time (in seconds); mean values and standard deviations. Notice
that most $\mathrm{SSE}$ axes have an overall scaling factor indicated
at the top. Each point of the evolutionary algorithms corresponds
to a different $J$. The same data, along with the values of $J$
used, is reported in table~\ref{tab:results2}. Bottom-right panel:
evolution of the average $\mathrm{SSE}$ through the generations for
the \emph{Miss America} dataset, using the largest values of $J$
(the endpoints of the curves correspond to the right-most points in
the bottom-right panel). Error bars are small and were omitted for
clarity. The generation-$0$ value for \noun{ga-kmeans-raw} is not
shown because it is $\left(895\pm15\right)\times10^{3}$.}
\end{figure*}

\begin{table*}
\begin{centering}
\caption{\label{tab:results2}Real-world datasets: $\mathrm{SSE}$ costs vs
time, varying $J$}
\par\end{centering}
\begin{centering}
{\scriptsize{}}%
\begin{tabular}{|c|ccc|ccc|ccc|}
\hline 
\multirow{2}{*}{{\scriptsize{}dataset}} & \multicolumn{3}{c|}{\noun{\scriptsize{}recombinator}} & \multicolumn{3}{c|}{\noun{\scriptsize{}ga-raw}} & \multicolumn{3}{c|}{\noun{\scriptsize{}ga++}}\tabularnewline
 & {\scriptsize{}$J$} & {\scriptsize{}$t$} & {\scriptsize{}$\mathrm{SSE}$} & {\scriptsize{}$J$} & {\scriptsize{}$t$} & {\scriptsize{}$\mathrm{SSE}$} & {\scriptsize{}$J$} & {\scriptsize{}$t$} & {\scriptsize{}$\mathrm{SSE}$}\tabularnewline
\hline 
\hline 
\multirow{5}{*}{\emph{\scriptsize{}Bridge}} & {\scriptsize{}$5$} & {\scriptsize{}$2.59\pm0.22$} & {\scriptsize{}$10662\pm16$} & {\scriptsize{}$15$} & {\scriptsize{}$2.07\pm0.23$} & {\scriptsize{}$10619\pm10$} & {\scriptsize{}$15$} & {\scriptsize{}$2.46\pm0.24$} & {\scriptsize{}$10611\pm11$}\tabularnewline
 & {\scriptsize{}$10$} & {\scriptsize{}$7.6\pm0.5$} & {\scriptsize{}$10590\pm10$} & {\scriptsize{}$45$} & {\scriptsize{}$6.8\pm0.7$} & {\scriptsize{}$10589\pm12$} & {\scriptsize{}$45$} & {\scriptsize{}$7.7\pm0.8$} & {\scriptsize{}$10588\pm13$}\tabularnewline
 & {\scriptsize{}$20$} & {\scriptsize{}$20.4\pm1.3$} & {\scriptsize{}$10554\pm7$} & {\scriptsize{}$120$} & {\scriptsize{}$19.2\pm1.7$} & {\scriptsize{}$10570\pm11$} & {\scriptsize{}$120$} & {\scriptsize{}$22.3\pm1.6$} & {\scriptsize{}$10569\pm7$}\tabularnewline
 & {\scriptsize{}$40$} & {\scriptsize{}$59\pm4$} & {\scriptsize{}$10535\pm5$} & {\scriptsize{}$351$} & {\scriptsize{}$57\pm5$} & {\scriptsize{}$10558\pm8$} & {\scriptsize{}$351$} & {\scriptsize{}$68\pm4$} & {\scriptsize{}$10559\pm8$}\tabularnewline
 & {\scriptsize{}$80$} & {\scriptsize{}$181\pm11$} & {\scriptsize{}$10527\pm4$} & {\scriptsize{}$861$} & {\scriptsize{}$145\pm8$} & {\scriptsize{}$10549\pm7$} & {\scriptsize{}$861$} & {\scriptsize{}$170\pm9$} & {\scriptsize{}$105049\pm7$}\tabularnewline
\hline 
\multirow{5}{*}{\emph{\scriptsize{}House}} & {\scriptsize{}$5$} & {\scriptsize{}$16.5\pm1.9$} & {\scriptsize{}$935.4\pm1.2$} & {\scriptsize{}$10$} & {\scriptsize{}$13.1\pm1.3$} & {\scriptsize{}$934.2\pm0.9$} & {\scriptsize{}$10$} & {\scriptsize{}$14.5\pm1.4$} & {\scriptsize{}$933.7\pm0.7$}\tabularnewline
 & {\scriptsize{}$10$} & {\scriptsize{}$42\pm4$} & {\scriptsize{}$933.0\pm0.9$} & {\scriptsize{}$28$} & {\scriptsize{}$38\pm3$} & {\scriptsize{}$932.6\pm0.7$} & {\scriptsize{}$28$} & {\scriptsize{}$42\pm3$} & {\scriptsize{}$932.3\pm0.6$}\tabularnewline
 & {\scriptsize{}$20$} & {\scriptsize{}$106\pm8$} & {\scriptsize{}$931.4\pm0.5$} & {\scriptsize{}$66$} & {\scriptsize{}$94\pm8$} & {\scriptsize{}$931.7\pm0.6$} & {\scriptsize{}$66$} & {\scriptsize{}$104\pm7$} & {\scriptsize{}$931.3\pm0.5$}\tabularnewline
 & {\scriptsize{}$40$} & {\scriptsize{}$240\pm19$} & {\scriptsize{}$930.9\pm0.4$} & {\scriptsize{}$153$} & {\scriptsize{}$211\pm17$} & {\scriptsize{}$931.1\pm0.5$} & {\scriptsize{}$153$} & {\scriptsize{}$236\pm19$} & {\scriptsize{}$930.9\pm0.5$}\tabularnewline
 & {\scriptsize{}$80$} & {\scriptsize{}$553\pm48$} & {\scriptsize{}$930.4\pm0.3$} & {\scriptsize{}$378$} & {\scriptsize{}$533\pm39$} & {\scriptsize{}$930.5\pm0.3$} & {\scriptsize{}$378$} & {\scriptsize{}$592\pm29$} & {\scriptsize{}$930.4\pm0.3$}\tabularnewline
\hline 
\multirow{5}{*}{\emph{\scriptsize{}M. Am.}} & {\scriptsize{}$5$} & {\scriptsize{}$5.4\pm0.4$} & {\scriptsize{}$526.8\pm0.7$} & {\scriptsize{}$15$} & {\scriptsize{}$4.5\pm0.5$} & {\scriptsize{}$525.7\pm0.8$} & {\scriptsize{}$15$} & {\scriptsize{}$4.7\pm0.5$} & {\scriptsize{}$524.4\pm0.8$}\tabularnewline
 & {\scriptsize{}$10$} & {\scriptsize{}$15.7\pm0.9$} & {\scriptsize{}$523.6\pm0.5$} & {\scriptsize{}$45$} & {\scriptsize{}$15.2\pm1.3$} & {\scriptsize{}$523.9\pm0.8$} & {\scriptsize{}$45$} & {\scriptsize{}$15.4\pm1.4$} & {\scriptsize{}$523.1\pm0.4$}\tabularnewline
 & {\scriptsize{}$20$} & {\scriptsize{}$41.2\pm2.0$} & {\scriptsize{}$521.8\pm0.4$} & {\scriptsize{}$120$} & {\scriptsize{}$40\pm4$} & {\scriptsize{}$522.9\pm0.6$} & {\scriptsize{}$120$} & {\scriptsize{}$43\pm5$} & {\scriptsize{}$522.4\pm0.5$}\tabularnewline
 & {\scriptsize{}$40$} & {\scriptsize{}$112\pm7$} & {\scriptsize{}$520.9\pm0.3$} & {\scriptsize{}$276$} & {\scriptsize{}$95\pm8$} & {\scriptsize{}$522.2\pm0.5$} & {\scriptsize{}$276$} & {\scriptsize{}$103\pm8$} & {\scriptsize{}$521.8\pm0.4$}\tabularnewline
 & {\scriptsize{}$80$} & {\scriptsize{}$324\pm26$} & {\scriptsize{}$520.3\pm0.2$} & {\scriptsize{}$741$} & {\scriptsize{}$266\pm22$} & {\scriptsize{}$521.6\pm0.3$} & {\scriptsize{}$741$} & {\scriptsize{}$272\pm25$} & {\scriptsize{}$521.5\pm0.5$}\tabularnewline
\hline 
\multirow{3}{*}{\emph{\scriptsize{}Urb.GB}} & {\scriptsize{}$5$} & {\scriptsize{}$299\pm39$} & {\scriptsize{}$229.3\pm0.4$} & {\scriptsize{}$21$} & {\scriptsize{}$234\pm29$} & {\scriptsize{}$233.5\pm2.4$} & {\scriptsize{}$21$} & {\scriptsize{}$292\pm30$} & {\scriptsize{}$228.82\pm0.19$}\tabularnewline
 & {\scriptsize{}$10$} & {\scriptsize{}$820\pm60$} & {\scriptsize{}$228.42\pm0.19$} & {\scriptsize{}$55$} & {\scriptsize{}$600\pm90$} & {\scriptsize{}$231.5\pm1.6$} & {\scriptsize{}$55$} & {\scriptsize{}$770\pm70$} & {\scriptsize{}$228.64\pm0.17$}\tabularnewline
 & {\scriptsize{}$20$} & {\scriptsize{}$2000\pm180$} & {\scriptsize{}$228.11\pm0.13$} & {\scriptsize{}$136$} & {\scriptsize{}$1530\pm190$} & {\scriptsize{}$230.2\pm0.7$} & {\scriptsize{}$136$} & {\scriptsize{}$1970\pm230$} & {\scriptsize{}$228.45\pm0.11$}\tabularnewline
\hline 
\multirow{5}{*}{\emph{\scriptsize{}Olivetti}} & {\scriptsize{}$5$} & {\scriptsize{}$5.7\pm0.6$} & {\scriptsize{}$11595\pm33$} & {\scriptsize{}$21$} & {\scriptsize{}$4.3\pm0.6$} & {\scriptsize{}$11542\pm11$} & {\scriptsize{}$21$} & {\scriptsize{}$5.7\pm0.5$} & {\scriptsize{}$11538\pm10$}\tabularnewline
 & {\scriptsize{}$10$} & {\scriptsize{}$15.7\pm1.7$} & {\scriptsize{}$11538\pm14$} & {\scriptsize{}$55$} & {\scriptsize{}$10.8\pm1.3$} & {\scriptsize{}$11531\pm9$} & {\scriptsize{}$55$} & {\scriptsize{}$14.7\pm1.4$} & {\scriptsize{}$11528\pm8$}\tabularnewline
 & {\scriptsize{}$20$} & {\scriptsize{}$41\pm4$} & {\scriptsize{}$11522\ensuremath{\pm8}$} & {\scriptsize{}$153$} & {\scriptsize{}$29\pm3$} & {\scriptsize{}$11524\pm6$} & {\scriptsize{}$153$} & {\scriptsize{}$39.6\pm2.7$} & {\scriptsize{}$11521\pm7$}\tabularnewline
 & {\scriptsize{}$40$} & {\scriptsize{}$118\pm15$} & {\scriptsize{}$11513\pm6$} & {\scriptsize{}$435$} & {\scriptsize{}$79\pm6$} & {\scriptsize{}$11518\pm7$} & {\scriptsize{}$435$} & {\scriptsize{}$111\pm6$} & {\scriptsize{}$11516\pm6$}\tabularnewline
 & {\scriptsize{}$80$} & {\scriptsize{}$355\pm54$} & {\scriptsize{}$11510\pm4$} & {\scriptsize{}$1225$} & {\scriptsize{}$209\pm17$} & {\scriptsize{}$11514\pm6$} & {\scriptsize{}$1225$} & {\scriptsize{}$307\pm16$} & {\scriptsize{}$11510\pm5$}\tabularnewline
\hline 
\end{tabular}{\scriptsize\par}
\par\end{centering}
\centering{}\medskip{}
Results for the tests on real-world datasets (evolutionary algorithms
only). The same data is shown in the first five panels of fig.~\ref{fig:results2}.
The $\mathrm{SSE}$s for \emph{Bridge}, \emph{House} and \emph{Miss
America }have been scaled down by a factor of $10^{3}$.
\end{table*}

The results are summarized in fig.~\ref{fig:results2}, and the same
data (excluding \noun{randswap-kmeans++}) is also reported in table~\ref{tab:results2},
along with the values of $J$ that were used. We can preliminarily
observe that the three evolutionary algorithms are superior by a wide
margin to \noun{randswap-kmeans++} in all cases except one in which
good seeding is crucial \noun{(ga-kmeans-raw} on \emph{UrbanGB}),
confirming the effectiveness of the evolutionary approach for hard
optimization challenges. We also observe that in all cases the curves
for \noun{ga-kmeans++} are uniformly better or equal than the curves
for \noun{ga-kmeans-raw}. This suggests to always prefer greedy-$k$-means++
seeding over naive random sampling in difficult scenarios. When comparing
\noun{ga-kmeans++} with \noun{recombinator-kmeans}, we find that in
all cases at short times \noun{ga-kmeans++} is a clear winner. At
longer times, though, \noun{recombinator-kmeans} always catches up,
and in three cases out of five it overcomes \noun{ga-kmeans++}. These
claims are confirmed by a detailed statistical significance analysis,
reported in Appendix~\ref{appendix:significance}. This suggests
a time-cost trade-off: \noun{ga-kmeans++} should be preferred when
the time available for the optimization is shorter, and \noun{recombinator-kmeans}
when attaining a better $\mathrm{SSE}$ is more important. Another
concern could be memory usage, as maintaining large populations can
be expensive, especially if $k$ is large. Indeed, the memory requirements
in both algorithms are dominated by the storage of the configurations,
each of which requires $kD+N$ floating point numbers (for the centroids
$\mathcal{C}$ and and auxiliary structures) and $N+k$ integers (for
the partition $\mathcal{P}$ and auxiliary structures). In principle
though \noun{ga-kmeans} can be implemented using only $O\left(\sqrt{J}\right)$
storage: assuming $J=J_{e}\left(J_{e}-1\right)/2$ we can simply keep
only the best $J_{e}$ configurations even as we build the new generation
(although a parallel implementation achieving this would not be trivial).
Taking this into account, \noun{ga-kmeans} is better than \noun{recombinator-kmeans}
(which uses all $J$ configurations instead) in this regard, by a
factor between $1$ and $2$.

The bottom-right panel in fig.~\ref{fig:results2} shows the evolution
of the $\mathrm{SSE}$ cost of evolutionary algorithms as the generations
progress, in a representative case. The curves for all the other experiments
of this section are qualitatively very similar. The starting values
(generation $0$) reflect the statistics of the seeding: we can observe
that greedy-$k$-means++ followed by local optimization, while clearly
superior to the ``raw'' initialization, is unable to achieve the
results of the population algorithms, not even by repeated restarts
(the population algorithms' final results are more than $16$ standard
deviations below). The general evolution indicates that the PNN crossover
scheme is greedier and achieves lower costs faster, and converges
in fewer iterations, whereas the \noun{reservoir-kmeans++} crossover
requires more generations but can in some cases find better solutions.
It is worth noting that, due to the elitist policy, the genetic algorithm
builds each new population of $J=741$ individuals by using only the
information in the $J_{e}=39$ elite ones, roughly half of the $J=80$
which are used in \noun{reservoir-kmeans++} recombination. The strategy
thus pays off in the initial generations and leads to bigger gains.
The soft-selection mechanism of \noun{recombinator-kmeans}, based
on sampling from the entire population with a progressively stronger
bias, produces slower gains at the beginning and takes longer to converge,
but in the end it seems to make equal or better use of the initial
pool of individuals, at least for large populations.

\begin{table*}
\begin{centering}
\caption{\label{tab:stats2}Real-world datasets: additional statistics}
{\scriptsize{}}%
\begin{tabular}{|c|cccc|cccc|cccc|}
\hline 
\multirow{2}{*}{{\scriptsize{}dataset}} & \multicolumn{4}{c|}{\noun{\scriptsize{}recombinator}} & \multicolumn{4}{c|}{\noun{\scriptsize{}ga-raw}} & \multicolumn{4}{c|}{\noun{\scriptsize{}ga++}}\tabularnewline
 & {\scriptsize{}$J$} & {\scriptsize{}$n_{\mathrm{gen}}$} & {\scriptsize{}$n_{\mathrm{Ll}}$} & {\scriptsize{}$t_{\times}$} & {\scriptsize{}$J$} & {\scriptsize{}$n_{\mathrm{gen}}$} & {\scriptsize{}$n_{\mathrm{Ll}}$} & {\scriptsize{}$t_{\times}$} & {\scriptsize{}$J$} & {\scriptsize{}$n_{\mathrm{gen}}$} & {\scriptsize{}$n_{\mathrm{Ll}}$} & {\scriptsize{}$t_{\times}$}\tabularnewline
\hline 
\hline 
\multirow{5}{*}{\emph{\scriptsize{}Bridge}} & {\scriptsize{}$5$} & {\scriptsize{}$11.4\pm1.1$} & {\scriptsize{}$0.38\pm0.03$} & {\scriptsize{}$86\%$} & {\scriptsize{}$15$} & {\scriptsize{}$11.8\pm2.1$} & {\scriptsize{}$1.04\pm0.10$} & {\scriptsize{}$54\%$} & {\scriptsize{}$15$} & {\scriptsize{}$10.9\pm2.4$} & {\scriptsize{}$0.90\pm0.13$} & {\scriptsize{}$58\%$}\tabularnewline
 & {\scriptsize{}$10$} & {\scriptsize{}$16.0\pm1.2$} & {\scriptsize{}$1.04\pm0.06$} & {\scriptsize{}$87\%$} & {\scriptsize{}$45$} & {\scriptsize{}$13.3\pm2.3$} & {\scriptsize{}$3.4\pm0.3$} & {\scriptsize{}$55\%$} & {\scriptsize{}$45$} & {\scriptsize{}$11.5\pm2.6$} & {\scriptsize{}$2.9\pm0.4$} & {\scriptsize{}$58\%$}\tabularnewline
 & {\scriptsize{}$20$} & {\scriptsize{}$19.8\pm1.3$} & {\scriptsize{}$2.5\pm0.1$} & {\scriptsize{}$89\%$} & {\scriptsize{}$120$} & {\scriptsize{}$13.9\pm2.2$} & {\scriptsize{}$9.6\pm0.8$} & {\scriptsize{}$55\%$} & {\scriptsize{}$120$} & {\scriptsize{}$13.0\pm1.9$} & {\scriptsize{}$8.4\pm0.8$} & {\scriptsize{}$59\%$}\tabularnewline
 & {\scriptsize{}$40$} & {\scriptsize{}$22.9\pm1.4$} & {\scriptsize{}$5.7\pm0.3$} & {\scriptsize{}$91\%$} & {\scriptsize{}$351$} & {\scriptsize{}$14.6\pm2.0$} & {\scriptsize{}$28.2\pm2.4$} & {\scriptsize{}$56\%$} & {\scriptsize{}$351$} & {\scriptsize{}$14.0\pm1.7$} & {\scriptsize{}$25.8\pm2.0$} & {\scriptsize{}$59\%$}\tabularnewline
 & {\scriptsize{}$80$} & {\scriptsize{}$25.9\pm1.5$} & {\scriptsize{}$12.7\pm0.6$} & {\scriptsize{}$94\%$} & {\scriptsize{}$861$} & {\scriptsize{}$14.9\pm1.4$} & {\scriptsize{}$72\pm4$} & {\scriptsize{}$56\%$} & {\scriptsize{}$861$} & {\scriptsize{}$14.5\pm1.6$} & {\scriptsize{}$65\pm5$} & {\scriptsize{}$59\%$}\tabularnewline
\hline 
\multirow{5}{*}{\emph{\scriptsize{}House}} & {\scriptsize{}$5$} & {\scriptsize{}$10.0\pm1.5$} & {\scriptsize{}$0.47\pm0.06$} & {\scriptsize{}$64\%$} & {\scriptsize{}$10$} & {\scriptsize{}$11.3\pm2.0$} & {\scriptsize{}$1.04\pm0.15$} & {\scriptsize{}$18\%$} & {\scriptsize{}$10$} & {\scriptsize{}$9.9\pm2.3$} & {\scriptsize{}$0.91\pm0.17$} & {\scriptsize{}$19\%$}\tabularnewline
 & {\scriptsize{}$10$} & {\scriptsize{}$13.1\pm1.3$} & {\scriptsize{}$1.2\pm0.1$} & {\scriptsize{}$65\%$} & {\scriptsize{}$28$} & {\scriptsize{}$11.9\pm2.3$} & {\scriptsize{}$3.0\pm0.4$} & {\scriptsize{}$18\%$} & {\scriptsize{}$28$} & {\scriptsize{}$10.1\pm1.7$} & {\scriptsize{}$2.6\pm0.3$} & {\scriptsize{}$19\%$}\tabularnewline
 & {\scriptsize{}$20$} & {\scriptsize{}$15.9\pm1.5$} & {\scriptsize{}$2.9\pm0.2$} & {\scriptsize{}$68\%$} & {\scriptsize{}$66$} & {\scriptsize{}$12.6\pm2.1$} & {\scriptsize{}$7.4\pm1.0$} & {\scriptsize{}$19\%$} & {\scriptsize{}$66$} & {\scriptsize{}$11.8\pm1.7$} & {\scriptsize{}$6.8\pm0.8$} & {\scriptsize{}$20\%$}\tabularnewline
 & {\scriptsize{}$40$} & {\scriptsize{}$17.8\pm1.7$} & {\scriptsize{}$6.4\pm0.5$} & {\scriptsize{}$70\%$} & {\scriptsize{}$153$} & {\scriptsize{}$12.1\pm1.6$} & {\scriptsize{}$16.7\pm1.9$} & {\scriptsize{}$18\%$} & {\scriptsize{}$153$} & {\scriptsize{}$11.4\pm2.3$} & {\scriptsize{}$15.3\pm2.4$} & {\scriptsize{}$20\%$}\tabularnewline
 & {\scriptsize{}$80$} & {\scriptsize{}$19.7\pm2.0$} & {\scriptsize{}$13.8\pm1.2$} & {\scriptsize{}$73\%$} & {\scriptsize{}$378$} & {\scriptsize{}$12.8\pm1.6$} & {\scriptsize{}$43\pm5$} & {\scriptsize{}$19\%$} & {\scriptsize{}$378$} & {\scriptsize{}$11.7\pm1.2$} & {\scriptsize{}$39\pm3$} & {\scriptsize{}$20\%$}\tabularnewline
\hline 
\multirow{5}{*}{\emph{\scriptsize{}M. Am.}} & {\scriptsize{}$5$} & {\scriptsize{}$13.4\pm1.2$} & {\scriptsize{}$0.59\pm0.04$} & {\scriptsize{}$74\%$} & {\scriptsize{}$15$} & {\scriptsize{}$12.0\pm2.5$} & {\scriptsize{}$1.34\pm0.14$} & {\scriptsize{}$36\%$} & {\scriptsize{}$15$} & {\scriptsize{}$9.5\pm2.5$} & {\scriptsize{}$1.10\pm0.17$} & {\scriptsize{}$36\%$}\tabularnewline
 & {\scriptsize{}$10$} & {\scriptsize{}$19.1\pm1.2$} & {\scriptsize{}$1.67\pm0.09$} & {\scriptsize{}$75\%$} & {\scriptsize{}$45$} & {\scriptsize{}$14.1\pm2.0$} & {\scriptsize{}$4.6\pm0.4$} & {\scriptsize{}$37\%$} & {\scriptsize{}$45$} & {\scriptsize{}$11.0\pm2.6$} & {\scriptsize{}$3.7\pm0.5$} & {\scriptsize{}$37\%$}\tabularnewline
 & {\scriptsize{}$20$} & {\scriptsize{}$24.3\pm1.4$} & {\scriptsize{}$4.24\pm0.21$} & {\scriptsize{}$76\%$} & {\scriptsize{}$120$} & {\scriptsize{}$13.9\pm2.5$} & {\scriptsize{}$12.3\pm1.2$} & {\scriptsize{}$37\%$} & {\scriptsize{}$120$} & {\scriptsize{}$11.9\pm2.9$} & {\scriptsize{}$10.6\pm1.6$} & {\scriptsize{}$38\%$}\tabularnewline
 & {\scriptsize{}$40$} & {\scriptsize{}$28.3\pm1.7$} & {\scriptsize{}$9.7\pm0.5$} & {\scriptsize{}$80\%$} & {\scriptsize{}$276$} & {\scriptsize{}$14.4\pm2.4$} & {\scriptsize{}$29.1\pm2.7$} & {\scriptsize{}$37\%$} & {\scriptsize{}$276$} & {\scriptsize{}$13.2\pm2.3$} & {\scriptsize{}$25.9\pm2.5$} & {\scriptsize{}$39\%$}\tabularnewline
 & {\scriptsize{}$80$} & {\scriptsize{}$31.6\pm2.0$} & {\scriptsize{}$21.9\pm1.3$} & {\scriptsize{}$84\%$} & {\scriptsize{}$741$} & {\scriptsize{}$15.2\pm2.2$} & {\scriptsize{}$82\pm7$} & {\scriptsize{}$37\%$} & {\scriptsize{}$741$} & {\scriptsize{}$12.6\pm2.6$} & {\scriptsize{}$69\pm9$} & {\scriptsize{}$37\%$}\tabularnewline
\hline 
\multirow{3}{*}{\emph{\scriptsize{}Urb.GB}} & {\scriptsize{}$5$} & {\scriptsize{}$9.8\pm1.6$} & {\scriptsize{}$0.39\pm0.04$} & {\scriptsize{}$84\%$} & {\scriptsize{}$21$} & {\scriptsize{}$7.8\pm1.8$} & {\scriptsize{}$1.5\pm0.3$} & {\scriptsize{}$20\%$} & {\scriptsize{}$21$} & {\scriptsize{}$5.0\pm1.4$} & {\scriptsize{}$1.01\pm0.26$} & {\scriptsize{}$21\%$}\tabularnewline
 & {\scriptsize{}$10$} & {\scriptsize{}$14.0\pm1.3$} & {\scriptsize{}$1.14\pm0.10$} & {\scriptsize{}$84\%$} & {\scriptsize{}$55$} & {\scriptsize{}$8.0\pm2.7$} & {\scriptsize{}$4.0\pm1.1$} & {\scriptsize{}$21\%$} & {\scriptsize{}$55$} & {\scriptsize{}$5.4\pm2.1$} & {\scriptsize{}$2.8\pm0.9$} & {\scriptsize{}$21\%$}\tabularnewline
 & {\scriptsize{}$20$} & {\scriptsize{}$17.3\pm1.7$} & {\scriptsize{}$2.79\pm0.28$} & {\scriptsize{}$84\%$} & {\scriptsize{}$136$} & {\scriptsize{}$8.1\pm2.5$} & {\scriptsize{}$10.1\pm2.2$} & {\scriptsize{}$20\%$} & {\scriptsize{}$136$} & {\scriptsize{}$5.8\pm1.8$} & {\scriptsize{}$7.4\pm2.1$} & {\scriptsize{}$22\%$}\tabularnewline
\hline 
\multirow{5}{*}{\emph{\scriptsize{}Olivetti}} & {\scriptsize{}$5$} & {\scriptsize{}$8.1\pm1.0$} & {\scriptsize{}$0.11\pm0.01$} & {\scriptsize{}$93\%$} & {\scriptsize{}$21$} & {\scriptsize{}$6.7\pm1.3$} & {\scriptsize{}$0.38\pm0.05$} & {\scriptsize{}$73\%$} & {\scriptsize{}$21$} & {\scriptsize{}$6.1\pm1.4$} & {\scriptsize{}$0.28\pm0.06$} & {\scriptsize{}$80\%$}\tabularnewline
 & {\scriptsize{}$10$} & {\scriptsize{}$10.4\pm1.3$} & {\scriptsize{}$0.29\pm0.03$} & {\scriptsize{}$94\%$} & {\scriptsize{}$55$} & {\scriptsize{}$6.6\pm1.2$} & {\scriptsize{}$0.95\pm0.11$} & {\scriptsize{}$73\%$} & {\scriptsize{}$55$} & {\scriptsize{}$5.7\pm1.1$} & {\scriptsize{}$0.67\pm0.13$} & {\scriptsize{}$80\%$}\tabularnewline
 & {\scriptsize{}$20$} & {\scriptsize{}$11.9\pm1.2$} & {\scriptsize{}$0.65\pm0.05$} & {\scriptsize{}$95\%$} & {\scriptsize{}$153$} & {\scriptsize{}$6.6\pm0.9$} & {\scriptsize{}$2.53\pm0.25$} & {\scriptsize{}$73\%$} & {\scriptsize{}$153$} & {\scriptsize{}$5.8\pm0.9$} & {\scriptsize{}$1.81\pm0.27$} & {\scriptsize{}$81\%$}\tabularnewline
 & {\scriptsize{}$40$} & {\scriptsize{}$13.4\pm1.8$} & {\scriptsize{}$1.45\pm0.16$} & {\scriptsize{}$96\%$} & {\scriptsize{}$435$} & {\scriptsize{}$6.5\pm0.8$} & {\scriptsize{}$7.0\pm0.6$} & {\scriptsize{}$73\%$} & {\scriptsize{}$435$} & {\scriptsize{}$5.5\pm0.7$} & {\scriptsize{}$4.7\pm0.5$} & {\scriptsize{}$81\%$}\tabularnewline
 & {\scriptsize{}$80$} & {\scriptsize{}$13.9\pm2.2$} & {\scriptsize{}$3.0\pm0.4$} & {\scriptsize{}$97\%$} & {\scriptsize{}$1225$} & {\scriptsize{}$6.2\pm0.7$} & {\scriptsize{}$18.9\pm1.4$} & {\scriptsize{}$73\%$} & {\scriptsize{}$1225$} & {\scriptsize{}$5.3\pm0.7$} & {\scriptsize{}$13.0\pm1.5$} & {\scriptsize{}$81\%$}\tabularnewline
\hline 
\end{tabular}{\scriptsize\par}
\par\end{centering}
\centering{}\medskip{}
Some statistics for the tests on real-world datasets, cf.~table~\ref{tab:results2}.
For each algorithm, the columns represent: $n_{\mathrm{gen}}$ = average
number of generations until convergence; $n_{\mathrm{Ll}}$ = average
number of Lloyd's iterations, in thousands, excluding generation $0$;
$t_{\times}$ = the fraction of wall-clock time spent in the recombination/crossover
process.
\end{table*}

In table~\ref{tab:stats2} we show additional statistics for these
tests, which generally confirm this picture. Indeed, we observe that
the number of generations required for convergence, $n_{\mathrm{gen}}$,
is generally higher for \noun{recombinator-kmeans}, and furthermore
that they tend to increase with $J$ for this algorithm, whereas they
tend to remain stable for \noun{ga-kmeans}. On average, \noun{ga-kmeans-raw}
requires roughly one or two additional generations to converge, compared
to \noun{ga-kmeans++}; in other words, the time saved to avoid careful
seeding is spent later in the process, and the trade-off is generally
not advantageous.

In the same table, we also report the average number of Lloyd's iterations
($n_{\mathrm{Ll}}$), excluding those used to produce generation $0$.
It turns out in all cases that $n_{\mathrm{Ll}}\approx t_{\mathrm{avg}}Jn_{\mathrm{gen}}$,
where $t_{\mathrm{avg}}<t_{\mathrm{max}}$ is dataset-dependent. It's
also worth pointing out that the wall-clock time spent for each individual
Lloyd iteration decreases nearly linearly towards zero as the generations
progress, because as the configurations converge and become more stable
a progressively larger fraction of the required distance computations
can be skipped. We also report the fraction of total time spent in
the recombination/crossover process ($t_{\times}$). We see that \noun{recombinator-kmeans}
spends considerably more time in the crossover, and correspondingly
that it performs fewer Lloyd's iterations in the given amount of time,
compared to \noun{ga-kmeans}; furthermore, $t_{\times}$ increases
with $J$ for \noun{recombinator-kmeans} whereas it stays stable for
\noun{ga-kmeans}; this is due to the fact that the time for the PNN
crossover also decreases at each generation as the configurations
stabilize, like it happens for Lloyd's iterations, while the time
for \noun{reservoir-kmeans++} stays basically constant.

\begin{table*}
\begin{centering}
\caption{\label{tab:VI}Real-world datasets: assessing the quality of the solutions}
{\scriptsize{}}%
\begin{tabular}{|c|cc|cc|cc|}
\hline 
\multirow{2}{*}{{\scriptsize{}dataset}} & \multicolumn{2}{c|}{\noun{\scriptsize{}recombinator}} & \multicolumn{2}{c|}{\noun{\scriptsize{}ga-raw}} & \multicolumn{2}{c|}{\noun{\scriptsize{}ga++}}\tabularnewline
 & {\scriptsize{}$J$} & {\scriptsize{}$\mathrm{VI}$} & {\scriptsize{}$J$} & {\scriptsize{}$\mathrm{VI}$} & {\scriptsize{}$J$} & {\scriptsize{}$\mathrm{VI}$}\tabularnewline
\hline 
\hline 
\multirow{3}{*}{\emph{\scriptsize{}Urb.GB}} & {\scriptsize{}$5$} & {\scriptsize{}$1.829\pm0.009$} & {\scriptsize{}$21$} & {\scriptsize{}$1.875\pm0.09$} & {\scriptsize{}$21$} & {\scriptsize{}$1.854\pm0.004$}\tabularnewline
 & {\scriptsize{}$10$} & {\scriptsize{}$1.828\pm0.006$} & {\scriptsize{}$55$} & {\scriptsize{}$1.865\pm0.06$} & {\scriptsize{}$55$} & {\scriptsize{}$1.851\pm0.004$}\tabularnewline
 & {\scriptsize{}$20$} & {\scriptsize{}$1.829\pm0.007$} & {\scriptsize{}$136$} & {\scriptsize{}$1.861\pm0.05$} & {\scriptsize{}$136$} & {\scriptsize{}$1.851\pm0.005$}\tabularnewline
\hline 
\multirow{5}{*}{\emph{\scriptsize{}Olivetti}} & {\scriptsize{}$5$} & {\scriptsize{}$1.48\pm0.05$} & {\scriptsize{}$21$} & {\scriptsize{}$1.45\pm0.03$} & {\scriptsize{}$21$} & {\scriptsize{}$1.45\pm0.03$}\tabularnewline
 & {\scriptsize{}$10$} & {\scriptsize{}$1.47\pm0.03$} & {\scriptsize{}$55$} & {\scriptsize{}$1.45\pm0.03$} & {\scriptsize{}$55$} & {\scriptsize{}$1.44\pm0.03$}\tabularnewline
 & {\scriptsize{}$20$} & {\scriptsize{}$1.45\pm0.02$} & {\scriptsize{}$153$} & {\scriptsize{}$1.45\pm0.03$} & {\scriptsize{}$153$} & {\scriptsize{}$1.43\pm0.03$}\tabularnewline
 & {\scriptsize{}$40$} & {\scriptsize{}$1.45\pm0.02$} & {\scriptsize{}$435$} & {\scriptsize{}$1.44\pm0.03$} & {\scriptsize{}$435$} & {\scriptsize{}$1.43\pm0.03$}\tabularnewline
 & {\scriptsize{}$80$} & {\scriptsize{}$1.45\pm0.01$} & {\scriptsize{}$1225$} & {\scriptsize{}$1.44\pm0.03$} & {\scriptsize{}$1225$} & {\scriptsize{}$1.43\pm0.01$}\tabularnewline
\hline 
\end{tabular}{\scriptsize\par}
\par\end{centering}
\centering{}\medskip{}
Variation of Information (VI) between the solutions and the ground-truth
partitions, for the \emph{UrbanGB} and \emph{Olivetti} datasets.
\end{table*}

Finally, we report in table~\ref{tab:VI} the results of the $\mathrm{VI}$
analysis for the \emph{UrbanGB} and \emph{Olivetti} datasets. The
dependency on $J$ appears to be very mild, except possibly for the
smallest population sizes. For the \emph{Olivetti} dataset, the results
are rather similar between the algorithms, with perhaps a small advantage
for \noun{ga-kmeans}. For \emph{UrbanGB}, on the other hand, there
is a clear ranking: \noun{recombinator-kmeans} is better than \noun{ga-kmeans++}
which is better than \noun{ga-kmeans-raw}. Part of it might be explained
by the better $\mathrm{SSE}$ costs achieved by \noun{recombinator-kmeans}
and \noun{ga-kmeans++} compared to \noun{ga-kmeans-raw}. However,
even when the costs between \noun{recombinator-kmeans} and \noun{ga-keans++}
are comparable, the former achieves a better VI. This suggests the
existence of a non-negligible algorithm-dependent bias in the distribution
of the final configurations, even when the costs are indistinguishable.
Obtaining a better characterization of this phenomenon would be very
useful for the design of improved algorithms.

\section{Discussion\label{sec:discussion}}

We have investigated and contrasted two evolutionary algorithms for
minimum-sum-of-squares optimization, on a variety of challenging datasets:
a novel algorithm based on a whole-population recombination process
and a variable selection-for-variation mechanism, \noun{recombinator-kmeans},
and an efficient and augmented implementation of an existing genetic
algorithm with a specialized pairwise crossover and an elitist selection-for-survival
scheme, \noun{ga-kmeans}.

Both algorithms are rather efficient, can be parallelized very easily,
and produce uniformly good results, although the original \noun{ga-kmeans}
suffers from poor initialization quality in a few cases. Indeed, one
of our findings was to show the benefit of augmenting \noun{ga-kmeans}
with greedy-$k$-means++ seeding. In challenging scenarios, the resulting
algorithm always offers the best time-cost trade-off for relatively
short times/small population sizes; and even in modestly complicated
artificial cases it's arguably the best overall choice, also compared
to simple local-search algorithms. \noun{ga-kmeans} is also efficient
in terms of memory required, due to its elitist selection that allows
to discard part of the population entirely.

On the other hand, \noun{recombinator-kmeans} is still competitive
at short times and can produce better results in the long run. Its
crossover mechanism is overall more costly and requires more generations
for convergence, but the reweighted stochastic recombination scheme
seems to be able to better exploit the initial population, since it
gives better results (or equal, if the global optimum is hit) for
a given population size. In the case of the \emph{UrbanGB} dataset,
which is the largest and arguably most challenging dataset that we
tested, we even found that it is able to find a better approximation
of the ground truth. We should note here that we performed some experiments
in which we applied an elitist selection policy with the \noun{reservoir-kmeans++}
crossover, and they turned out not to be competitive with the other
algorithms, which indicates that the progressively-stronger selection-for-variation
is an important component of \noun{recombinator-kmeans.} Presumably,
this is because the weighting scheme is applied on top of the existing
probability distribution already employed by $k$-means++, which may
thus overcome a small prior and pick good centroids even from otherwise
relatively bad configurations.

The \noun{recombinator-kmeans} scheme has a few additional advantages.
The main one in our opinion is that it is rather general and not as
closely tied to the minimum-sum-of-squares problem as \noun{ga-kmeans}.
Since $k$-means++ initialization can be adapted to other clustering
problems, like $k$-medians or $k$-medoids, so can the \noun{reservoir-kmeans++}
recombination; preliminary tests on these cases have indeed produced
very good results (with the caveat that for $k$-medoids optimization
at least some approximate version of the triangle inequality is necessary
for $k$-means++ to make sense and perform well). It should even be
possible to improve it further by straightforwardly borrowing from
any proposed way to speed-up the $k$-means++ procedure, such as that
of ref.~\citep{bachem2016fast} (which is based on using a Monte
Carlo Markov Chain in order to perform the sampling and is thus fully
compatible with \noun{reservoir-means++}). Another appealing quality
of the scheme is its simplicity: the code is only marginally more
complicated to implement than multi-start-$k$-means with $k$-means++
initialization (which is itself quite standard and very simple), since
it largely reuses the same algorithms and data structures, only framing
it in a population-based iterative algorithm.

Striving for simplicity, we did not include mutation mechanisms in
our tests. As a consequence, both algorithms have to rely only on
the initial population as a source of variability. On top of this,
the \noun{ga-kmeans} algorithm uses a deterministic selection policy
and crossover, and \noun{recombinator-kmeans} employs an explicit
mechanism to force the population to collapse. Indeed, our results
seem to suggest that for this problem it is more convenient to improve
the size or the quality of the initial population, by increasing $J$
and using greedy-$k$-means++ seeding, than adding random mutations.
It is still possible that a better mutation scheme, potentially in
combination with an adaptive population size scheme \citep{lobo2005review},
could allow to achieve the same results with smaller populations.

More generally, our results confirm the benefit of population algorithms
compared to local search ones for clustering applications, even when
careful initialization is used and the possibility of multiple restarts
is accounted for. Our proposed scheme may also be of more general
interest in the broader context of evolutionary and population algorithms
because of its peculiar features. Its multi-parent crossover is certainly
tailored to representative-based clustering problems, but in its essence
it stochastically pieces together different locally-optimal configurations
in a relatively simple way, exploiting prior knowledge (encoded in
the seeding algorithm) about the kind of configurations that are likely
to be good for the optimization problem. The progressively stronger
biasing mechanism, used in place of hard selection, shifting from
exploration to exploitation and ensuring convergence, is also an original
contribution. Both of these features could in principle find wider
applicability than clustering.

\section*{Acknowledgments}

I wish to thank R. Zecchina for interesting discussions and comments.
This work was supported by ONR Grant N00014-17-1-2569.

\newpage{}

\onecolumn

\appendix

\subsection{\label{appendix:case-study}A case-study analysis of the recombination
mechanism}

In this section, we present arguments and data in support of the heuristic
intuition at the base of recombinator-$k$-means, i.e.~using greedy-$k$-means++
as a recombination/crossover mechanism.

Preliminarily, we note that greedy-$k$-means++ seeding often leads
to much better results than uniform seeding, although it still offers
no guarantees and performing multiple restarts is standard practice
when possible. Under these circumstances, the extensive tests performed
recently by Celebi et al. \citep{celebi2013comparative} on an array
of on real and synthetic datasets comparing a large number of alternative
initialization methods concludes that in general greedy-$k$-means++
is arguably the best choice among the ones currently available, on
par with the method by Bradely and Fayyad \citep{bradley1998refining}.
The results of Fränti and Sieranoja \citep{franti2019much} seem to
indicate that the Maxmin seeding algorithm (which can be regarded
as a less-stochastic version of $k$-means++) would generally be preferable,
but their tests only use the non-greedy version of $k$-means++ seeding
(which can have significant impact). Note that Maxmin cannot be extended
in a greedy fashion like $k$-means++. Their results with the method
of Bradley and Fayyad are generally worse than even non-greedy $k$-means++.\footnote{It should be noted, however, that the survey of ref.~\citep{franti2019much}
is targeted towards obtaining a deeper understanding of how each seeding
algorithm (and Lloyd's algorithm in general) is affected by the properties
of the datasets, and thus they only use synthetic datasets. Ref.~\citep{celebi2013comparative},
on the other hand, tries to provide some guidelines for practitioners,
and test a large variety of real datasets as well.}

\begin{figure}
\centering{}\includegraphics[width=0.4\columnwidth]{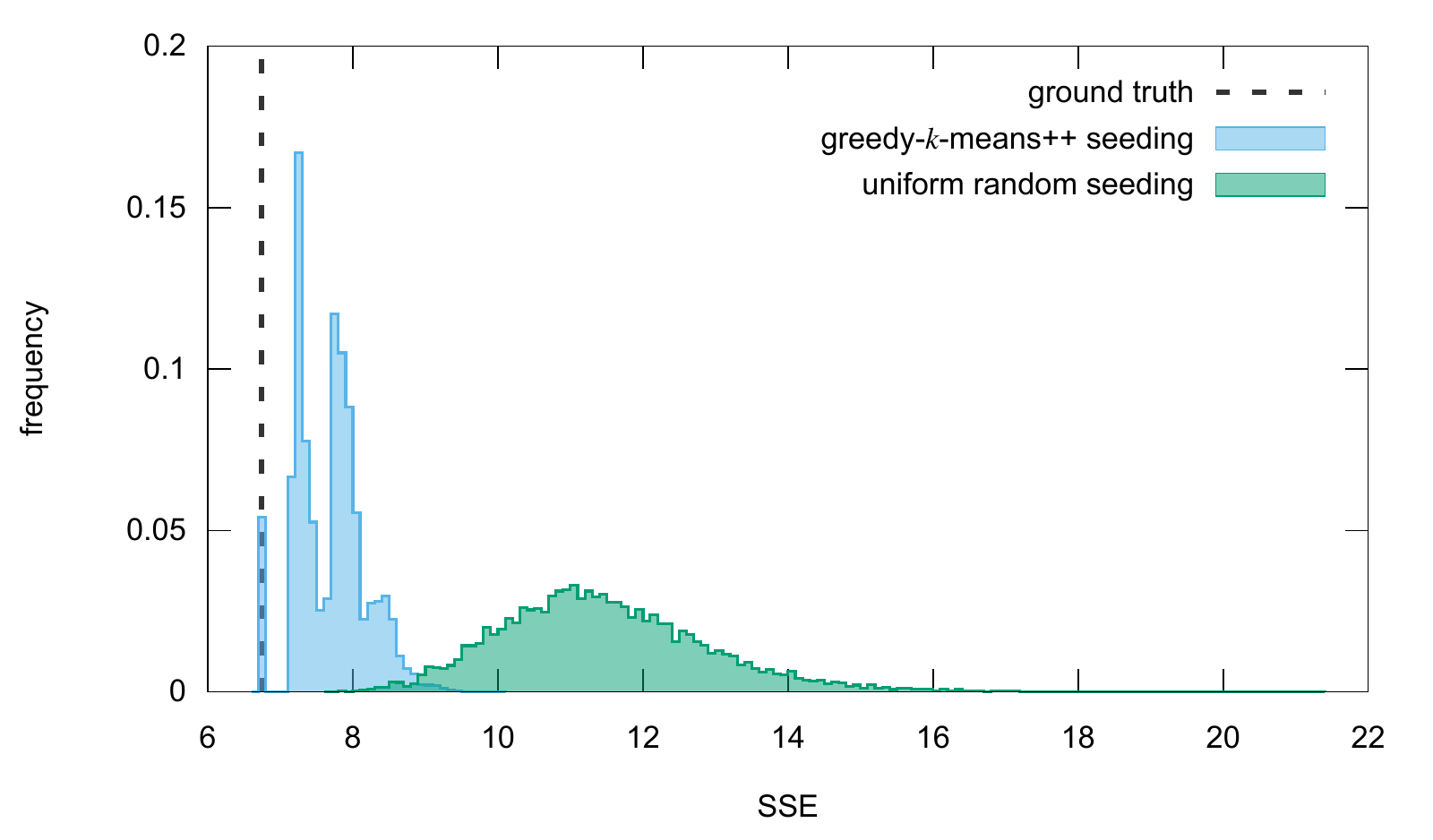}\caption{\label{fig:hist-A3}Histogram of the cost obtained from $10^{4}$
runs of Lloyd's algorithm with uniform initialization and with greedy-$k$-means++
initialization, on the \emph{A3} synthetic dataset (see main text;
we scaled the data uniformly to make it fit into a $\left[0,1\right]\times\text{\ensuremath{\left[0,1\right]}}$
square, by dividing all data entries by the overall maximum). The
latter algorithm clearly finds better configurations on average, and
in $5.4\%$ of the cases reaches the level of the ground truth (see
the first, isolated peak on the left). The following two peaks are,
roughly speaking, due to configurations with one or two \textquotedblleft mistakes\textquotedblright ,
respectively (see fig.~\ref{fig:examples-A3}).}
\end{figure}

\begin{figure*}
\begin{centering}
\includegraphics[width=1\textwidth]{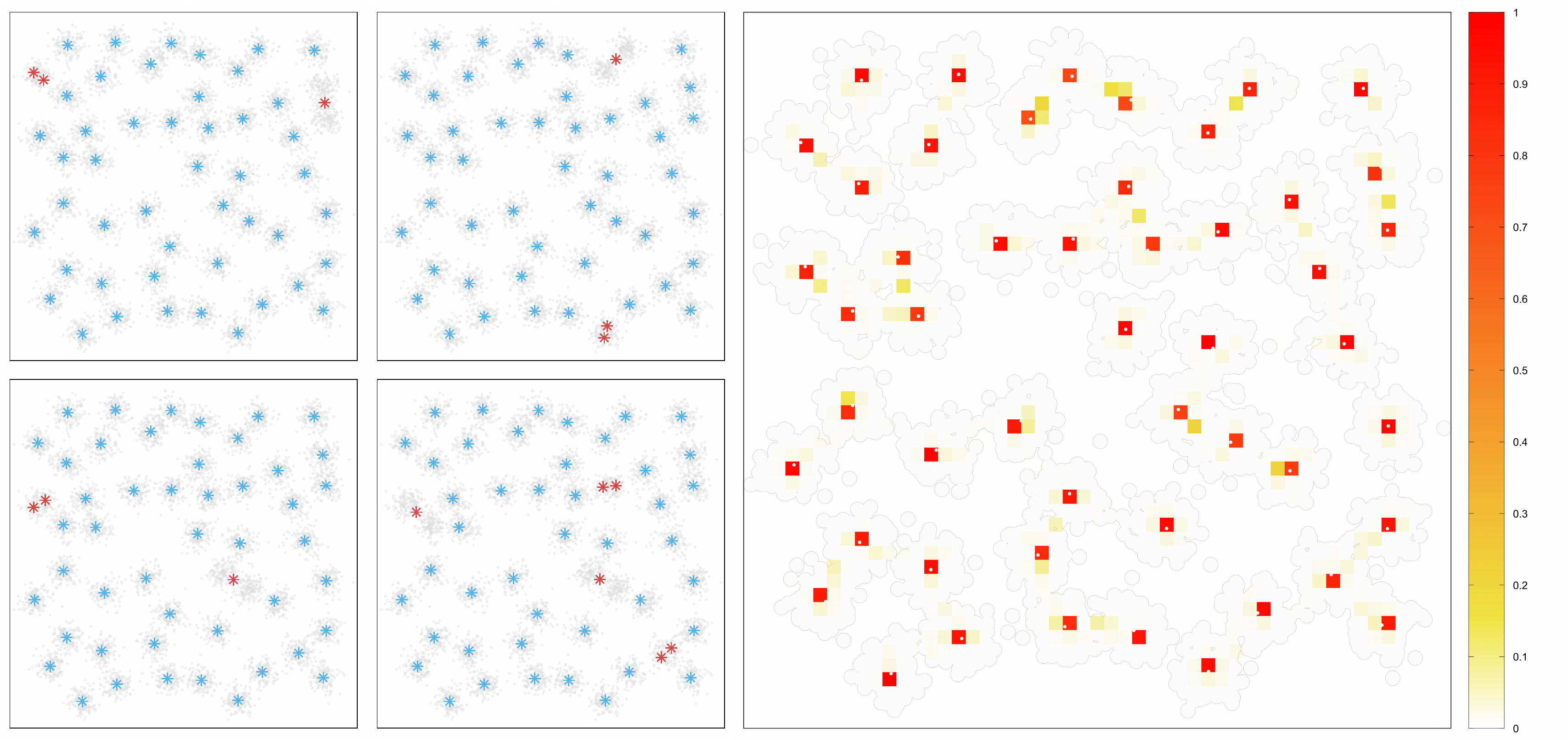}
\par\end{centering}
\caption{\label{fig:examples-A3}\emph{Left four panels.} Examples of \textquotedblleft typical\textquotedblright{}
configurations found by 4 independent runs of Lloyd's algorithm with
greedy-$k$-means++ seeding, on the $2$-dimensional \emph{A3} dataset
(same as for fig.~\ref{fig:hist-A3}). The data is represented in
gray, and the centroids in blue, except for \textquotedblleft mistakes\textquotedblright{}
which are shown in red. The first three panels show configurations
corresponding to the highest peak in the histogram of fig.~\ref{fig:hist-A3}
(one mistake each, centroid index $\mathrm{CI}=1$), while the bottom-right
panel shows a configuration corresponding to the following peak (two
mistakes, $\mathrm{CI}=2$). The fact that these mistakes are few
and mostly independent is shown quantitatively in the \emph{right
panel}, which depicts a two-dimensional histogram of the distribution
of the centroids found over $10^{4}$ runs by the local optimizer,
same data as for fig.~\ref{fig:hist-A3}. This graph aims at showing
the effect of superimposing several solutions like the ones shown
in the left panels. The area (a square of size $1$) was binned in
squares of size $0.02$. All of the red peaks correspond to the optimal
configuration (the optimal centroids are shown as small white dots
and they are indistinguishable from the ground truth at this level
of detail), and they all have a value of $0.7$ or more. Outside of
the optimal bins, all other bins are barely distinguishable from the
white background: none of them is higher than $0.22$, meaning that
no other location has a higher chance than $22\%$ of being produced
in a run of the algorithm. Overall, this shows that in examples like
this one the centroids of completed runs cluster extremely well around
the optimal ones.}
\end{figure*}

Here, we start by analyzing the results of multiple restarts on the
\emph{A3} synthetic dataset from the repository of ref.~\citep{franti2018k},
which is a moderately large two-dimensional dataset ($N=7500$, $D=2$)
consisting of $k=50$ fairly well-distinct homogeneous clusters. The
choice of this dataset is for illustrative purposes: it is designed
such that optimizing the $\mathrm{SSE}$ objective leads to a near-perfect
clustering, and yet finding such solution from scratch is a non-trivial
task. (For this dataset, the ground truth -- the centroids used for
generating the data -- is known, and the globally optimal configuration
is indeed extremely close to it.) Furthermore, it is two-dimensional,
and thus it can be easily visualized. Finally, as we shall show, it
is quite easy to visually identify and understand ``mistakes'' in
the clustering for the sub-optimal fixed points that are sufficiently
close to the global optimum.

We performed $10^{4}$ runs of local optimization (Lloyd's algorithm)
on this dataset, seeding the algorithm with greedy-$k$-means++, and
also as many runs with random uniform seeding (what's commonly referred
to as $k$-means) for comparison. The histogram of the costs is shown
in fig.~\ref{fig:hist-A3}. It can be noted that the uniform-seeding
version never reached the level of the ground truth cost. Greedy-$k$-means++
seeding on the other hand led to finding a cost comparable to that
of the ground truth in $5.4\%$ of the cases. The clear gap between
the ground-truth-level and the rest of the histogram allows us to
informally denote the configurations in the leftmost peak as ``successes'':
these are all configurations that any sensible method (from visual
inspection to more detailed analytical tools) would classify as near-perfect.
More formally, these configurations all have a \emph{centroid index}
(CI), as defined in ref.~\citep{franti2014centroid} , of $0$ (and
all other configurations, from the second peak onward, have $\mathrm{CI>}0$).
With the observed percentage of success of $5.4\%$ of greedy-$k$-means++,
we can easily compute (with the inverse of a geometric cumulative
distribution function) that one would need about $R=54$ runs to reach
a confidence level of $95\%$ of success with a simple multiple restarts
strategy, and $R=83$ to reach a confidence of $99\%$.

A further inspection of fig.~\ref{fig:hist-A3} shows that, even
when unsuccessful, the algorithm has a few clearly visible peaks that
are close to the optimum. These peaks correspond to situations that
are depicted in the four left panels of figure~\ref{fig:examples-A3}.
Overall, one can qualitatively observe that, in most runs, most centroids
end up being near their optimal position, with only a few, fairly
well identifiable ``mistakes''. Moreover, these mistakes are mostly
independent from one run to the next. This is confirmed when superimposing
the final centroids obtained from different runs, shown in the right
panel of figure~\ref{fig:examples-A3}, whereby we see how they form
tight, dense clusters around the optimal centroids.

\subsection{\label{appendix:parameters-deltab}Parameters choice: The effect
of varying $\Delta\beta$}

\begin{table*}
\begin{centering}
{\scriptsize{}}%
\begin{tabular}{|c|cc|cc|cc|}
\hline 
\multirow{2}{*}{{\scriptsize{}$\Delta\beta$}} & \multicolumn{2}{c|}{\emph{\scriptsize{}A3}} & \multicolumn{2}{c|}{\emph{\scriptsize{}Birch1}} & \multicolumn{2}{c|}{\emph{\scriptsize{}Birch2}}\tabularnewline
 & {\scriptsize{}time ($s$)} & {\scriptsize{}succ.rate} & {\scriptsize{}time ($s$)} & {\scriptsize{}succ.rate} & {\scriptsize{}time ($s$)} & {\scriptsize{}succ.rate}\tabularnewline
\hline 
\hline 
{\scriptsize{}$0$} & {\scriptsize{}$0.091\pm0.016$} & {\scriptsize{}$100\%$} & {\scriptsize{}$3.9\pm0.3$} & {\scriptsize{}$100\%$} & {\scriptsize{}$2.2\pm0.4$} & {\scriptsize{}$100\%$}\tabularnewline
{\scriptsize{}$0.1$} & {\scriptsize{}$0.089\pm0.014$} & {\scriptsize{}$100\%$} & {\scriptsize{}$4.0\pm0.3$} & {\scriptsize{}$100\%$} & {\scriptsize{}$2.2\pm0.4$} & {\scriptsize{}$100\%$}\tabularnewline
{\scriptsize{}$1$} & {\scriptsize{}$0.084\pm0.014$} & {\scriptsize{}$99.9\%$} & {\scriptsize{}$3.7\pm0.3$} & {\scriptsize{}$100\%$} & {\scriptsize{}$2.1\pm0.3$} & {\scriptsize{}$100\%$}\tabularnewline
{\scriptsize{}$10$} & {\scriptsize{}$0.079\pm0.012$} & {\scriptsize{}$97.3\%$} & {\scriptsize{}$3.6\pm0.3$} & {\scriptsize{}$93\%$} & {\scriptsize{}$1.9\pm0.2$} & {\scriptsize{}$100\%$}\tabularnewline
\hline 
\end{tabular}{\scriptsize\par}
\par\end{centering}
\begin{centering}
\medskip{}
\par\end{centering}
\caption{\label{tab:vary_db}Average and standard deviation of convergence
time (in seconds), and success rate, for \noun{recombinator-kmeans}
on\emph{ }a few synthetic datasets, varying $\Delta\beta$, $1000$
samples per point for \emph{A3} and $100$ samples per point for \emph{Birch1}
and \emph{Birch2}.}
\end{table*}
\begin{figure*}
\begin{centering}
\includegraphics[width=1\columnwidth]{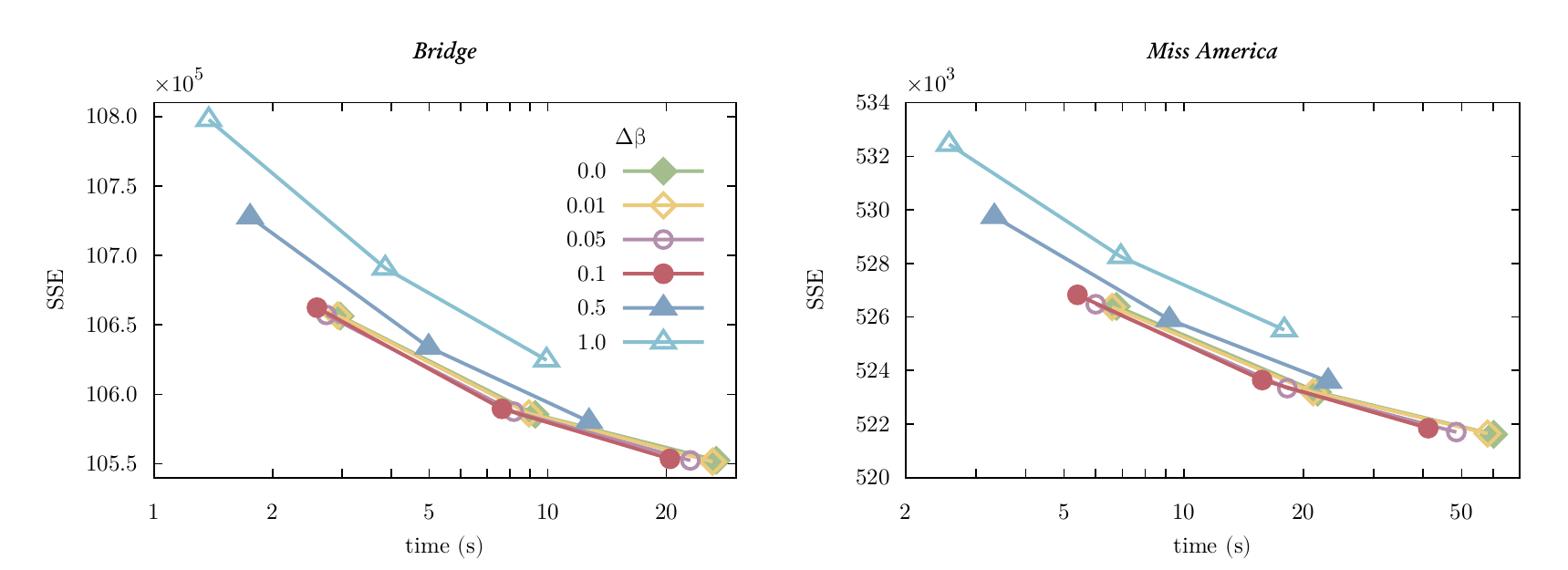}
\par\end{centering}
\caption{\label{fig:vary_db}Results of \noun{recombinator-kmeans} on two datasets,
\emph{Bridge} and \emph{Miss America}, with different $\Delta\beta$
and varying $J$ in $\left\{ 5,10,20\right\} $; $500$ samples per
point (average values shown, error bars omitted for clarity).}
\end{figure*}

In table~\ref{tab:vary_db} we show the average convergence time
and success rate of \noun{recombinator-kmeans} with $J=5$ and different
$\Delta\beta$, for three synthetic datasets. This is an extension
of the result shown for $\Delta\beta=0.1$ in table~2 of the main
text. The $\Delta\beta=0$ case corresponds to uniform weighting.
For these datasets (and similar ones with a well-defined clustering
structure, like those of sec.~VI.B of the main text) the performance
is only mildly affected by $\Delta\beta$ in a wide range; the success
rate is $100\%$ even with uniform weights and only starts to be affected
at very large $\Delta\beta$ (and in those cases a larger $J$ would
restore the perfect score). Indeed, in all these cases, the second
batch is already very likely to contain a ground-truth-level configuration,
as per the discussion in sec.~5 of the main text and sec.~\ref{appendix:case-study}
above; a third batch (or very rarely a fourth) may be required in
the harder cases only to detect convergence.

The situation is less straightforward when the optimization problems
are more challenging with no clear solution, as for the datasets of
sec.~VI.C of the main text. Figure~\ref{fig:vary_db} shows two
examples of varying the $\Delta\beta$ parameter of \noun{recombinator-kmeans},
on the \emph{Bridge} and \emph{Miss America} datasets. Our tests with
other datasets and configurations suggest that the behavior shown
is fairly well representative of challenging problems. A few exceptions
manifested with very small values of $\Delta\beta$: setting $\Delta\beta=0$
effectively removes the weighting and therefore the guarantee of convergence.
We have observed cases with less than 50\% chance of convergence and
major effects on convergence time with that setting.

From the figure we can observe that, in general, increasing $\Delta\beta$
reduces the convergence time (which is expected) but results in a
greedier algorithm. On balance, the setting $\Delta\beta=0.1$ seems
to be close to optimal in most cases. But the data also shows that
overall the effect of $\Delta\beta$ is rather mild.

\subsection{\label{appendix:parameters-tmax}Parameters choice: The effect of
varying $t_{\mathrm{max}}$}

\begin{figure*}
\begin{centering}
\includegraphics[width=1\columnwidth]{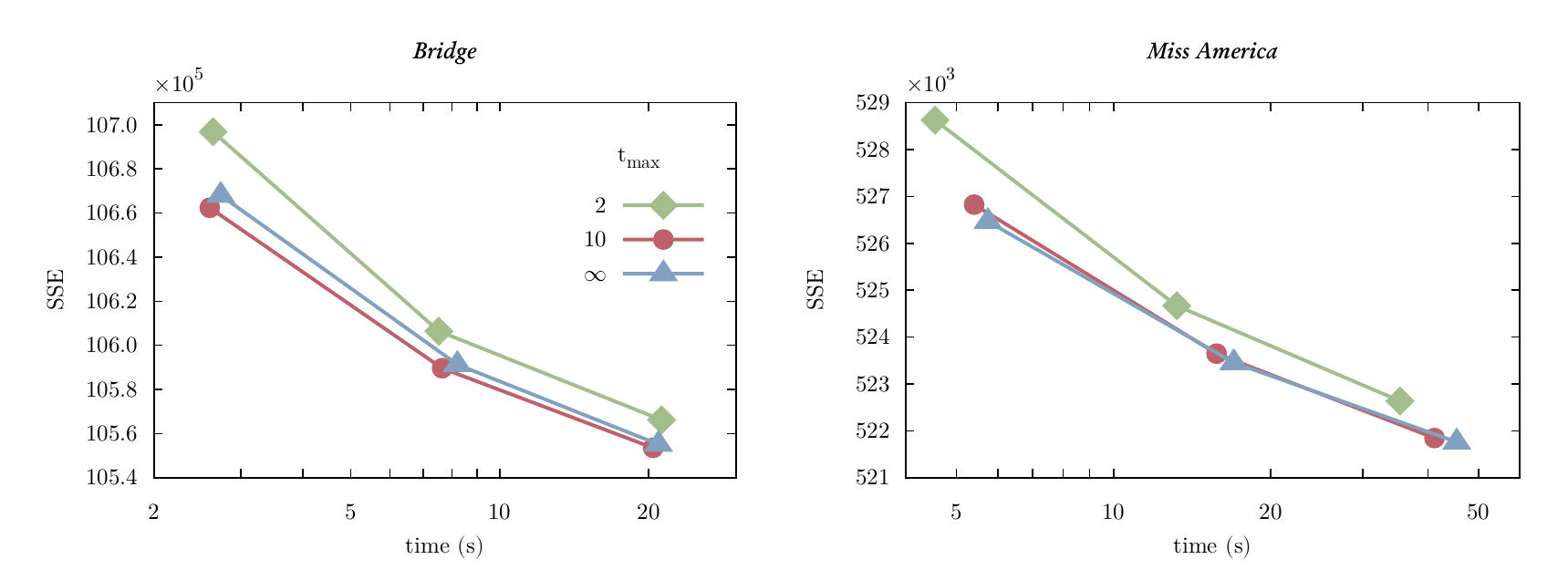}
\par\end{centering}
\caption{\label{fig:vary_tmax}Results of \noun{recombinator-kmeans} on two
datasets, \emph{Bridge} and \emph{Miss America}, with different $t_{\mathrm{max}}$
and varying $J$ in $\left\{ 5,10,20\right\} $; $500$ samples per
point (average values shown, error bars omitted for clarity).}
\end{figure*}

The effect of the $t_{\max}$ parameter for \noun{recombinator-kmeans}
depends on the dataset. For the synthetic datasets of the first batch
of tests, section~VI.B of the main text, using $t_{\max}=10$ instead
of allowing unlimited iterations has at most negligible effects, since
typically Lloyd's algorithm converges much earlier than the cutoff
in those cases (when seeded with greedy-$k$-means++).

In fig.~\ref{fig:vary_tmax} we show the effect of varying $t_{\max}$
in some more challenging cases, the \emph{Bridge} and \emph{Miss America}
datasets (see sec.~VI.C of the main text). In both cases we tested
$t_{\max}=2$, $10$ and unbounded. From these examples, which appear
to be fairly representative, it seems that capping the iterations
at $10$ can lead to some small improvements in the cost-vs-time tradeoff.
This appears to be the case even though the time gains of the cutoff
are diminished by the use of the technique of ref.~\citep{kaukoranta1999reduced}
that speeds up Lloyd iterations, and which is particularly effective
as the iterations progress. The setting $t_{\max}=2$, on the other
hand, appears to be excessive. Overall, as for the $\Delta\beta$
parameter, the effect of varying $t_{\mathrm{max}}$ appears to be
mild in a reasonably wide range.

\subsection{\label{appendix:significance}Statistical significance analysis}

In this section we report the results of a statistical significance
analysis on the results of sec.~VI.C of the main text. For each dataset,
we compared the $\mathrm{SSE}$ costs of pairs of different algorithms
at ``corresponding'' values of $J$: we compared \noun{ga-kmeans-raw}
with \noun{ga-kmeans++} at the same $J$; we compared \noun{recombinator-kmeans}
with \noun{ga-kmeans++} at values of $J$ that resulted in roughly
the same convergence time (see table~III of the main text); we compared
the largest available $J$ of \noun{recombinator-kmeans} with \noun{randswap-kmeans++}.

Each comparison was performed via two different non-parametric techniques.
Given two sets of costs, $\Phi_{a}$ and $\Phi_{b}$, possibly of
different sizes, we aimed at testing the null hypothesis that they
were originating from the same distribution. We used both a standard
two-sided Wilcoxon rank-sum test (also known as Mann-Whitney U test),
and a simple bootstrapping scheme, which we defined as follows. Denoting
with $\bar{\Phi}_{a}$ and $\bar{\Phi}_{b}$ the means of the costs
sets, we estimated the frequency, under the null hypotheses, with
which their distance $\left|\bar{\Phi}_{a}-\bar{\Phi}_{b}\right|$
would be larger or equal than the one measured. For this purpose,
we pooled together all the costs, $\Phi_{U}=\Phi_{a}\cup\Phi_{b}$,
and repeated $n_{B}=10^{5}$ times the following procedure: we randomly
reassigned the costs in $\Phi_{U}$ to two subsets with the same size
as the originals, and computed the distance of their means. We then
used the fraction of cases in which the distance exceeded the measured
one as the $p$-value.

This comparison is adequate for the purposes of determining whether
an algorithm is actually significantly better than another in terms
of the cost-vs-time trade-off only when the times are very similar.
This is generally the case for the \noun{recombinator }vs \noun{ga++}
and \noun{recombinator} vs \noun{randswap++} comparisons. However
for the \noun{ga-raw} vs \noun{ga++} comparison this test does not
adequately take into account the fact that \noun{ga-raw} is generally
slightly faster.
\begin{center}
{\scriptsize{}}
\begin{table}
\begin{centering}
{\scriptsize{}}%
\begin{tabular}{|c|ccc|ccc|ccc|}
\hline 
\multirow{2}{*}{{\scriptsize{}dataset}} & \multicolumn{3}{c|}{\noun{\scriptsize{}ga++}{\scriptsize{} vs}\noun{\scriptsize{} ga-raw}} & \multicolumn{3}{c|}{\noun{\scriptsize{}rec.}{\scriptsize{} vs }\noun{\scriptsize{}ga++}} & \multicolumn{3}{c|}{\noun{\scriptsize{}rec.}{\scriptsize{} vs }\noun{\scriptsize{}rs++}}\tabularnewline
 & {\scriptsize{}$J$} & {\scriptsize{}$p$-value} & {\scriptsize{}w} & {\scriptsize{}$J$} & {\scriptsize{}$p$-value} & {\scriptsize{}w} & {\scriptsize{}$J$} & {\scriptsize{}$p$-value} & {\scriptsize{}w}\tabularnewline
\hline 
\hline 
\multirow{5}{*}{\emph{\scriptsize{}Bridge}} & {\scriptsize{}$15$} & {\scriptsize{}$4.8\cdot10^{-3}$} & \textbf{\scriptsize{}2} & {\scriptsize{}$5$} & {\scriptsize{}$<10^{-4}$} & \textbf{\scriptsize{}2} &  &  & \tabularnewline
 & {\scriptsize{}$45$} & {\scriptsize{}$0.73$} & {\scriptsize{}2} & {\scriptsize{}$10$} & {\scriptsize{}$0.34$} & {\scriptsize{}2} &  &  & \tabularnewline
 & {\scriptsize{}$120$} & {\scriptsize{}$0.59$} & {\scriptsize{}2} & {\scriptsize{}$20$} & {\scriptsize{}$<10^{-4}$} & \textbf{\scriptsize{}1} &  &  & \tabularnewline
 & {\scriptsize{}$351$} & {\scriptsize{}$0.91$} & {\scriptsize{}3} & {\scriptsize{}$40$} & {\scriptsize{}$<10^{-4}$} & \textbf{\scriptsize{}1} &  &  & \tabularnewline
 & {\scriptsize{}$861$} & {\scriptsize{}$0.99$} & {\scriptsize{}3} & {\scriptsize{}$80$} & {\scriptsize{}$<10^{-4}$} & \textbf{\scriptsize{}1} & {\scriptsize{}$80$} & {\scriptsize{}$<10^{-4}$} & \textbf{\scriptsize{}1}\tabularnewline
\hline 
\multirow{5}{*}{\emph{\scriptsize{}House}} & {\scriptsize{}$10$} & {\scriptsize{}$1.4\cdot10^{-3}$} & \textbf{\scriptsize{}2} & {\scriptsize{}$5$} & {\scriptsize{}$<10^{-4}$} & \textbf{\scriptsize{}2} &  &  & \tabularnewline
 & {\scriptsize{}$28$} & {\scriptsize{}$6.8\cdot10^{-3}$} & \textbf{\scriptsize{}2} & {\scriptsize{}$10$} & {\scriptsize{}$<10^{-4}$} & \textbf{\scriptsize{}2} &  &  & \tabularnewline
 & {\scriptsize{}$66$} & {\scriptsize{}$5\cdot10^{-4}$} & \textbf{\scriptsize{}2} & {\scriptsize{}$20$} & {\scriptsize{}$0.34$} & {\scriptsize{}2} &  &  & \tabularnewline
 & {\scriptsize{}$153$} & {\scriptsize{}$5.2\cdot10^{-2}$} & {\scriptsize{}2} & {\scriptsize{}$40$} & {\scriptsize{}$0.88$} & {\scriptsize{}2} &  &  & \tabularnewline
 & {\scriptsize{}$378$} & {\scriptsize{}$2.4\cdot10^{-2}$} & {\scriptsize{}2} & {\scriptsize{}$80$} & {\scriptsize{}$0.87$} & {\scriptsize{}2} & {\scriptsize{}$80$} & {\scriptsize{}$<10^{-4}$} & \textbf{\scriptsize{}1}\tabularnewline
\hline 
\multirow{5}{*}{\emph{\scriptsize{}M. Am.}} & {\scriptsize{}$15$} & {\scriptsize{}$<10^{-4}$} & \textbf{\scriptsize{}2} & {\scriptsize{}$5$} & {\scriptsize{}$<10^{-4}$} & \textbf{\scriptsize{}2} &  &  & \tabularnewline
 & {\scriptsize{}$45$} & {\scriptsize{}$<10^{-4}$} & \textbf{\scriptsize{}2} & {\scriptsize{}$10$} & {\scriptsize{}$<10^{-4}$} & \textbf{\scriptsize{}2} &  &  & \tabularnewline
 & {\scriptsize{}$120$} & {\scriptsize{}$<10^{-4}$} & \textbf{\scriptsize{}2} & {\scriptsize{}$20$} & {\scriptsize{}$<10^{-4}$} & \textbf{\scriptsize{}1} &  &  & \tabularnewline
 & {\scriptsize{}$276$} & {\scriptsize{}$2\cdot10^{-4}$} & \textbf{\scriptsize{}2} & {\scriptsize{}$40$} & {\scriptsize{}$<10^{-4}$} & \textbf{\scriptsize{}1} &  &  & \tabularnewline
 & {\scriptsize{}$741$} & {\scriptsize{}$0.25$} & {\scriptsize{}2} & {\scriptsize{}$80$} & {\scriptsize{}$<10^{-4}$} & \textbf{\scriptsize{}1} & {\scriptsize{}$80$} & {\scriptsize{}$<10^{-4}$} & \textbf{\scriptsize{}1}\tabularnewline
\hline 
\multirow{3}{*}{\emph{\scriptsize{}Urb.GB}} & {\scriptsize{}$21$} & {\scriptsize{}$<10^{-4}$} & \textbf{\scriptsize{}2} & {\scriptsize{}$5$} & {\scriptsize{}$<10^{-4}$} & \textbf{\scriptsize{}2} &  &  & \tabularnewline
 & {\scriptsize{}$55$} & {\scriptsize{}$<10^{-4}$} & \textbf{\scriptsize{}2} & {\scriptsize{}$10$} & {\scriptsize{}$<10^{-4}$} & \textbf{\scriptsize{}1} &  &  & \tabularnewline
 & {\scriptsize{}$136$} & {\scriptsize{}$<10^{-4}$} & \textbf{\scriptsize{}2} & {\scriptsize{}$20$} & {\scriptsize{}$<10^{-4}$} & \textbf{\scriptsize{}1} & {\scriptsize{}$20$} & {\scriptsize{}$<10^{-4}$} & \textbf{\scriptsize{}1}\tabularnewline
\hline 
\multirow{5}{*}{\emph{\scriptsize{}Olivetti}} & {\scriptsize{}$21$} & {\scriptsize{}$1.3\cdot10^{-2}$} & {\scriptsize{}2} & {\scriptsize{}$5$} & {\scriptsize{}$<10^{-4}$} & \textbf{\scriptsize{}2} &  &  & \tabularnewline
 & {\scriptsize{}$55$} & {\scriptsize{}$5.2\cdot10^{-2}$} & {\scriptsize{}2} & {\scriptsize{}$10$} & {\scriptsize{}$<10^{-4}$} & \textbf{\scriptsize{}2} &  &  & \tabularnewline
 & {\scriptsize{}$153$} & {\scriptsize{}$1.1\cdot10^{-3}$} & \textbf{\scriptsize{}2} & {\scriptsize{}$20$} & {\scriptsize{}$0.69$} & {\scriptsize{}2} &  &  & \tabularnewline
 & {\scriptsize{}$435$} & {\scriptsize{}$0.11$} & {\scriptsize{}2} & {\scriptsize{}$40$} & {\scriptsize{}$2.4\cdot10^{-3}$} & \textbf{\scriptsize{}1} &  &  & \tabularnewline
 & {\scriptsize{}$1225$} & {\scriptsize{}$<10^{-4}$} & \textbf{\scriptsize{}2} & {\scriptsize{}$80$} & {\scriptsize{}$0.48$} & {\scriptsize{}2} & {\scriptsize{}$80$} & {\scriptsize{}$<10^{-4}$} & \textbf{\scriptsize{}1}\tabularnewline
\hline 
\end{tabular}{\scriptsize\par}
\par\end{centering}
\begin{centering}
\medskip{}
\par\end{centering}
\caption{\label{tab:significance}Significance analysis results (bootstrapping
scheme). The \textquotedblleft w\textquotedblright{} columns indicate
which of the two algorithms gave the best average cost, where $1$=\noun{recombinator},
2=\noun{ga++}, 3=\noun{ga-raw}, 4=\noun{randswap++}. The cases in
which the $p$-value is smaller than $10^{-2}$ are marked in bold.
We do not report precisely values smaller than $10^{-4}$ due to the
limited resolution of our bootstrapping scheme. The values of $J$
in the middle column refer to the \noun{recombinator} algorithm, while
the corresponding value of $J$ for \noun{ga++} is that of the first
column.}
\end{table}
{\scriptsize\par}
\par\end{center}

The results of the analysis are reported in table~\ref{tab:significance}.
We report only the $p$-values for the bootstrapping comparison, as
they are generally slightly larger than the ones from the Wilcoxon
rank-sum tests; however, the two tests are generally in excellent
agreement, and they are in complete agreement about which results
are significant when using a significance threshold of $10^{-2}$
(i.e., the ``w'' columns would look identical). The last column
shows that \noun{recombinator-kmeans} is significantly superior to
\noun{randswap-kmeans++} in all cases. The middle column shows that
\noun{ga-kmeans++} is generally superior to \noun{recombinator-kmeans}
at small $J$, but that at larger $J$ \noun{recombinator-kmeans}
catches up and, in 3 cases out of 5, ends up being significantly better.
The first column shows that \noun{ga-kmeans++} gives better or equal
costs than \noun{ga-kmeans-raw}. Although (as mentioned above) this
is measured at fixed $J$ rather than at fixed amount of time spent,
visual comparison with fig.~1 of the main text shows that most of
these results are sensible, except for the ones for the \emph{Bridge}
and \emph{Olivetti} dataset in which the few detected significant
results are almost certainly spurious.

\begin{thebibliography}{10}

\bibitem{wu2008top}
Xindong Wu, Vipin Kumar, J~Ross Quinlan, Joydeep Ghosh, Qiang Yang, Hiroshi
  Motoda, Geoffrey~J McLachlan, Angus Ng, Bing Liu, S~Yu Philip, et~al.
\newblock Top 10 algorithms in data mining.
\newblock {\em Knowledge and information systems}, 14(1):1--37, 2008.
\newblock URL: \url{https://doi.org/10.1007/s10115-007-0114-2}.

\bibitem{berkhin2006survey}
Pavel Berkhin.
\newblock A survey of clustering data mining techniques.
\newblock In {\em Grouping multidimensional data}, pages 25--71. Springer,
  2006.
\newblock URL: \url{https://doi.org/10.1007/3-540-28349-8_2}.

\bibitem{lloyd1982least}
Stuart Lloyd.
\newblock Least squares quantization in pcm.
\newblock {\em IEEE transactions on information theory}, 28(2):129--137, 1982.
\newblock URL: \url{https://doi.org/10.1109/TIT.1982.1056489}.

\bibitem{aloise2009np}
Daniel Aloise, Amit Deshpande, Pierre Hansen, and Preyas Popat.
\newblock Np-hardness of euclidean sum-of-squares clustering.
\newblock {\em Machine learning}, 75(2):245--248, 2009.
\newblock URL: \url{https://doi.org/10.1007/s10994-009-5103-0}.

\bibitem{celebi2013comparative}
M~Emre Celebi, Hassan~A Kingravi, and Patricio~A Vela.
\newblock A comparative study of efficient initialization methods for the
  k-means clustering algorithm.
\newblock {\em Expert systems with applications}, 40(1):200--210, 2013.
\newblock URL: \url{https://doi.org/10.1016/j.eswa.2012.07.021}.

\bibitem{franti2019much}
Pasi Fr{\"a}nti and Sami Sieranoja.
\newblock How much k-means can be improved by using better initialization and
  repeats?
\newblock {\em Pattern Recognition}, 2019.
\newblock URL: \url{https://doi.org/10.1016/j.patcog.2019.04.014}.

\bibitem{macqueen1967some}
James MacQueen.
\newblock Some methods for classification and analysis of multivariate
  observations.
\newblock In {\em Proceedings of the fifth Berkeley symposium on mathematical
  statistics and probability, volume 1: statistics}, volume~1, pages 281--297.
  Oakland, CA, USA, University of California Press, 1967.
\newblock URL: \url{https://projecteuclid.org/euclid.bsmsp/1200512992}.

\bibitem{arthur2007k}
David Arthur and Sergei Vassilvitskii.
\newblock k-means++: The advantages of careful seeding.
\newblock In {\em Proceedings of the eighteenth annual ACM-SIAM symposium on
  Discrete algorithms}, pages 1027--1035. Society for Industrial and Applied
  Mathematics, 2007.
\newblock URL:
  \url{http://theory.stanford.edu/~sergei/papers/kMeansPP-soda.pdf}.

\bibitem{goldberg1989genetic}
David~E Goldberg.
\newblock {\em Genetic algorithms in search, optimization and machine
  learning}.
\newblock Addison-Wesley Longman Publishing, 1989.
\newblock URL: \url{https://doi.org/10.5860/choice.27-0936}.

\bibitem{holland1992adaptation}
John~Henry Holland et~al.
\newblock {\em Adaptation in natural and artificial systems: an introductory
  analysis with applications to biology, control, and artificial intelligence}.
\newblock MIT press, 1992.
\newblock URL: \url{https://ieeexplore.ieee.org/servlet/opac?bknumber=6267401}.

\bibitem{koza1994genetic}
John~R Koza.
\newblock Genetic programming as a means for programming computers by natural
  selection.
\newblock {\em Statistics and computing}, 4(2):87--112, 1994.
\newblock URL: \url{https://doi.org/10.1007/BF00175355}.

\bibitem{hruschka2009survey}
Eduardo~Raul Hruschka, Ricardo~JGB Campello, Alex~A Freitas, et~al.
\newblock A survey of evolutionary algorithms for clustering.
\newblock {\em IEEE Transactions on Systems, Man, and Cybernetics, Part C
  (Applications and Reviews)}, 39(2):133--155, 2009.
\newblock URL: \url{https://doi.org/10.1109/TSMCC.2008.2007252}.

\bibitem{franti2000genetic}
Pasi Fr{\"a}nti.
\newblock Genetic algorithm with deterministic crossover for vector
  quantization.
\newblock {\em Pattern Recognition Letters}, 21(1):61--68, 2000.
\newblock URL: \url{https://doi.org/10.1016/S0167-8655(99)00133-6}.

\bibitem{moscato1992memetic}
Pablo Moscato and Michael~G Norman.
\newblock A "memetic" approach for the traveling salesman problem
  implementation of a computational ecology for combinatorial optimization on
  message-passing systems.
\newblock In {\em In Proceedings of the International Conference on Parallel
  Computing and Transputer Applications}, volume~1, pages 177--186. IOS Press,
  1992.
\newblock URL:
  \url{http://citeseerx.ist.psu.edu/viewdoc/summary?doi=10.1.1.50.1940}.

\bibitem{merz2002clustering}
Peter Merz and Andreas Zell.
\newblock Clustering gene expression profiles with memetic algorithms.
\newblock In {\em International Conference on Parallel Problem Solving from
  Nature}, pages 811--820. Springer, 2002.
\newblock URL: \url{https://doi.org/10.1007/3-540-45712-7_78}.

\bibitem{zhang1996birch}
Tian Zhang, Raghu Ramakrishnan, and Miron Livny.
\newblock Birch: an efficient data clustering method for very large databases.
\newblock {\em ACM sigmod record}, 25(2):103--114, 1996.
\newblock URL: \url{https://doi.org/10.1145/235968.233324}.

\bibitem{gan2017k}
Guojun Gan and Michael Kwok-Po Ng.
\newblock K-means clustering with outlier removal.
\newblock {\em Pattern Recognition Letters}, 90:8--14, 2017.
\newblock URL: \url{https://doi.org/10.1016/j.patrec.2017.03.008}.

\bibitem{liu2017sparse}
Weiwei Liu, Xiaobo Shen, and Ivor~W Tsang.
\newblock Sparse embedded k-means clustering.
\newblock In I.~Guyon, U.~V. Luxburg, S.~Bengio, H.~Wallach, R.~Fergus,
  S.~Vishwanathan, and R.~Garnett, editors, {\em Proceedings of the 31st
  International Conference on Neural Information Processing Systems},
  volume~30, pages 3321--3329. Curran Associates, Inc., 2017.
\newblock URL:
  \url{https://papers.nips.cc/paper/2017/hash/3214a6d842cc69597f9edf26df552e43-Abstract.html}.

\bibitem{kivijarvi2003self}
Juha Kivij{\"a}rvi, Pasi Fr{\"a}nti, and Olli Nevalainen.
\newblock Self-adaptive genetic algorithm for clustering.
\newblock {\em Journal of Heuristics}, 9(2):113--129, 2003.
\newblock URL: \url{https://doi.org/10.1023/A:1022521428870}.

\bibitem{tsutsui1998study}
Shigeyoshi Tsutsui and Ashish Ghosh.
\newblock A study on the effect of multi-parent recombination in real coded
  genetic algorithms.
\newblock In {\em 1998 IEEE international conference on evolutionary
  computation proceedings. IEEE World Congress on Computational Intelligence
  (Cat. No. 98TH8360)}, pages 828--833. IEEE, 1998.
\newblock URL: \url{https://doi.org/10.1109/ICEC.1998.700159}.

\bibitem{eiben2003multiparent}
Agoston~E Eiben.
\newblock Multiparent recombination in evolutionary computing.
\newblock In {\em Advances in evolutionary computing}, pages 175--192.
  Springer, 2003.
\newblock URL: \url{https://doi.org/10.1007/978-3-642-18965-4_6}.

\bibitem{kaukoranta1999reduced}
Timo Kaukoranta, P~Franti, and Olli Nevalainen.
\newblock Reduced comparison search for the exact gla.
\newblock In {\em Proceedings DCC'99 Data Compression Conference (Cat. No.
  PR00096)}, pages 33--41. IEEE, 1999.
\newblock URL: \url{https://doi.org/10.1109/DCC.1999.755651}.

\bibitem{scikit-learn}
F.~Pedregosa, G.~Varoquaux, A.~Gramfort, V.~Michel, B.~Thirion, O.~Grisel,
  M.~Blondel, P.~Prettenhofer, R.~Weiss, V.~Dubourg, J.~Vanderplas, A.~Passos,
  D.~Cournapeau, M.~Brucher, M.~Perrot, and E.~Duchesnay.
\newblock Scikit-learn: Machine learning in {P}ython.
\newblock {\em Journal of Machine Learning Research}, 12:2825--2830, 2011.
\newblock URL: \url{http://www.jmlr.org/papers/v12/pedregosa11a.html}.

\bibitem{scikit-learn-site}
scikit-learn.
\newblock URL: \url{https://scikit-learn.org}.

\bibitem{RECcode}
Code repository for recombinator-$k$-means and our implementation of
  {GA}-$k$-means.
\newblock URL: \url{https://github.com/carlobaldassi/RecombinatorKMeans.jl}.

\bibitem{back2018evolutionary}
Thomas B{\"a}ck, David~B Fogel, and Zbigniew Michalewicz.
\newblock {\em Evolutionary computation 1: Basic algorithms and operators}.
\newblock CRC press, 2018.

\bibitem{GAcode}
URL: \url{https://www.uef.fi/web/machine-learning/software}.

\bibitem{franti2018efficiency}
Pasi Fr{\"a}nti.
\newblock Efficiency of random swap clustering.
\newblock {\em Journal of Big Data}, 5(1):13, 2018.
\newblock URL: \url{https://doi.org/10.1186/s40537-018-0122-y}.

\bibitem{franti2018k}
Pasi Fr{\"a}nti and Sami Sieranoja.
\newblock K-means properties on six clustering benchmark datasets.
\newblock {\em Applied Intelligence}, 48(12):4743--4759, 2018.
\newblock URL: \url{http://cs.uef.fi/sipu/datasets/}.

\bibitem{franti2014centroid}
Pasi Fr{\"a}nti, Mohammad Rezaei, and Qinpei Zhao.
\newblock Centroid index: cluster level similarity measure.
\newblock {\em Pattern Recognition}, 47(9):3034--3045, 2014.
\newblock URL: \url{https://doi.org/10.1016/j.patcog.2014.03.017}.

\bibitem{UCI-MachineLearningRepository}
Dheeru Dua and Casey Graff.
\newblock {UCI} machine learning repository, 2017.
\newblock URL: \url{http://archive.ics.uci.edu/ml}.

\bibitem{OlivettiData}
URL:
  \url{https://scikit-learn.org/stable/datasets/real_world.html#the-olivetti-faces-dataset}.

\bibitem{meilua2007comparing}
Marina Meil{\u{a}}.
\newblock Comparing clusterings--an information based distance.
\newblock {\em Journal of multivariate analysis}, 98(5):873--895, 2007.
\newblock URL: \url{https://doi.org/10.1016/j.jmva.2006.11.013}.

\bibitem{bachem2016fast}
Olivier Bachem, Mario Lucic, Hamed Hassani, and Andreas Krause.
\newblock Fast and provably good seedings for k-means.
\newblock In D.~D. Lee, M.~Sugiyama, U.~V. Luxburg, I.~Guyon, and R.~Garnett,
  editors, {\em Advances in Neural Information Processing Systems 29}, pages
  55--63. Curran Associates, Inc., 2016.
\newblock URL:
  \url{http://papers.nips.cc/paper/6478-fast-and-provably-good-seedings-for-k-means.pdf}.

\bibitem{lobo2005review}
Fernando~G Lobo and Cl{\'a}udio~F Lima.
\newblock A review of adaptive population sizing schemes in genetic algorithms.
\newblock In {\em Proceedings of the 7th annual workshop on Genetic and
  evolutionary computation}, pages 228--234, 2005.
\newblock URL: \url{https://doi.org/10.1145/1102256.1102310}.

\bibitem{bradley1998refining}
Paul~S Bradley and Usama~M Fayyad.
\newblock Refining initial points for k-means clustering.
\newblock In {\em ICML}, volume~98, pages 91--99. Citeseer, 1998.
\newblock URL:
  \url{http://citeseerx.ist.psu.edu/viewdoc/download?doi=10.1.1.50.8528&rep=rep1&type=pdf}.

\end{thebibliography}
\end{document}